\documentclass[11pt]{article}

\usepackage[final]{acl}

\usepackage{times}
\usepackage{latexsym}

\usepackage[T1]{fontenc}
\usepackage[utf8]{inputenc}

\usepackage{microtype}

\usepackage{inconsolata}
\usepackage{algorithm}
\usepackage{algorithmic}

\usepackage{newfloat}
\usepackage{listings}
\usepackage{booktabs}     % for \toprule, \midrule, \bottomrule
\usepackage{makecell}     % for \makecell
\usepackage{xcolor}       % for \textcolor
\usepackage{amssymb}
\usepackage{pifont}
\usepackage{multirow} % Added for table multirow support
\usepackage{xspace}
\usepackage{colortbl}

\usepackage[most,skins,theorems]{tcolorbox}
\definecolor{darkblue}{rgb}{0, 0, 0.5}
\usepackage{longtable}

\tcbset{
  aibox/.style={
    width=\linewidth,
    top=8pt,
    bottom=4pt,
    colback=blue!6!white,
    colframe=black,
    colbacktitle=black,
    enhanced,
    center,
    attach boxed title to top left={yshift=-0.1in,xshift=0.15in},
    boxed title style={boxrule=0pt,colframe=white,},
  }
}
\newcolumntype{C}[1]{>{\centering\let\newline\\\arraybackslash\hspace{0pt}}m{#1}}
\newtcolorbox{AIbox}[2][]{aibox,title=#2,#1}

\DeclareCaptionStyle{ruled}{labelfont=normalfont,labelsep=colon,strut=off} % DO NOT CHANGE THIS
\floatstyle{ruled}
\newfloat{listing}{tb}{lst}{}
\floatname{listing}{Listing}

\usepackage{graphicx}

\newcommand{\system}{{MAS-Bench}\xspace}

\title{\system: A Unified Benchmark for Shortcut-Augmented \\ Hybrid Mobile GUI Agents}

\author{Pengxiang Zhao$^{1}$\thanks{Equal contribution.},
Guangyi Liu$^{1}$\footnotemark[1],
YaoZhen Liang$^{1}$\footnotemark[1],
Weiqing He$^{1}$,
Zhengxi Lu$^{1}$,\\
\textbf{WenHao Wang$^{1}$,
Yuehao Huang$^{1}$,
Yuxiang Chai$^{2}$,
Zhaolu Kang$^{3}$,
Yaxuan Guo$^{2}$,}\\
\textbf{Hao Wang$^{2}$,
Kexin Zhang$^{1}$\thanks{Corresponding author.},
Liang Liu$^{2}$\thanks{Project Lead.},
Yong Liu$^{1}$\footnotemark[2]}
\\[0.5em]
$^1$Zhejiang University
$^2$vivo AI Lab
$^3$Peking University
}

\usepackage{bibentry}
\usepackage{pifont}
\newcommand{\checkmarkmy}{\ding{51}}
\newcommand{\crossmy}{\ding{55}}

\begin{document}
\maketitle

\begin{abstract}
    Shortcuts such as APIs and deep-links have emerged as efficient complements to flexible GUI operations, fostering a promising hybrid paradigm for MLLM-based mobile automation. However, systematic evaluation of GUI–shortcut hybrid agents remains largely underexplored.
    To bridge this gap, we introduce \textbf{MAS-Bench}, a benchmark that pioneers the evaluation of GUI-shortcut hybrid agents with a specific focus on the mobile domain. Beyond merely using predefined shortcuts, MAS-Bench assesses an agent's capability to \textit{autonomously generate} shortcuts by discovering and creating reusable, low-cost workflows. It features 139 complex tasks across 11 real-world applications, a knowledge base of 88 predefined shortcuts (APIs, deep-links, RPA scripts), and 9 evaluation metrics.
    Experiments demonstrate that hybrid agents achieve up to 68.3\% success rate and 39\% greater execution efficiency than GUI-only counterparts. Furthermore, our evaluation framework effectively reveals the quality gap between predefined and agent-generated shortcuts, validating its capability to assess shortcut generation methods. MAS-Bench addresses the lack of systematic benchmarks for GUI-shortcut hybrid mobile agents, providing a foundational platform for future advancements in creating more efficient and robust intelligent agents. Project page: \url{https://pengxiang-zhao.github.io/MAS-Bench}.

\end{abstract}

\section{Introduction}

The rise of Large Language Models (LLMs)~\cite{openai2024gpt4technicalreport, huang2025cogddn, jiang2025think,li2026no} is driving the development of Graphical User Interface (GUI) Agents, enabling them to operate diverse digital platforms such as computers, web browsers, and smartphones~\cite{xu2024androidlab, liu2025llm, liu2025learnact}. Early research on mobile agents primarily focused on replicating human-like flexibility through GUI-only interaction~\cite{wang2025think,lu2026ui,xiao2026ui,lu2025uis1}. This approach grants agents the generality to operate on any application, but it often overlooks the significant efficiency advantages offered by more direct methods.

\begin{figure}[t]
  \centering
  \includegraphics[width=\columnwidth]{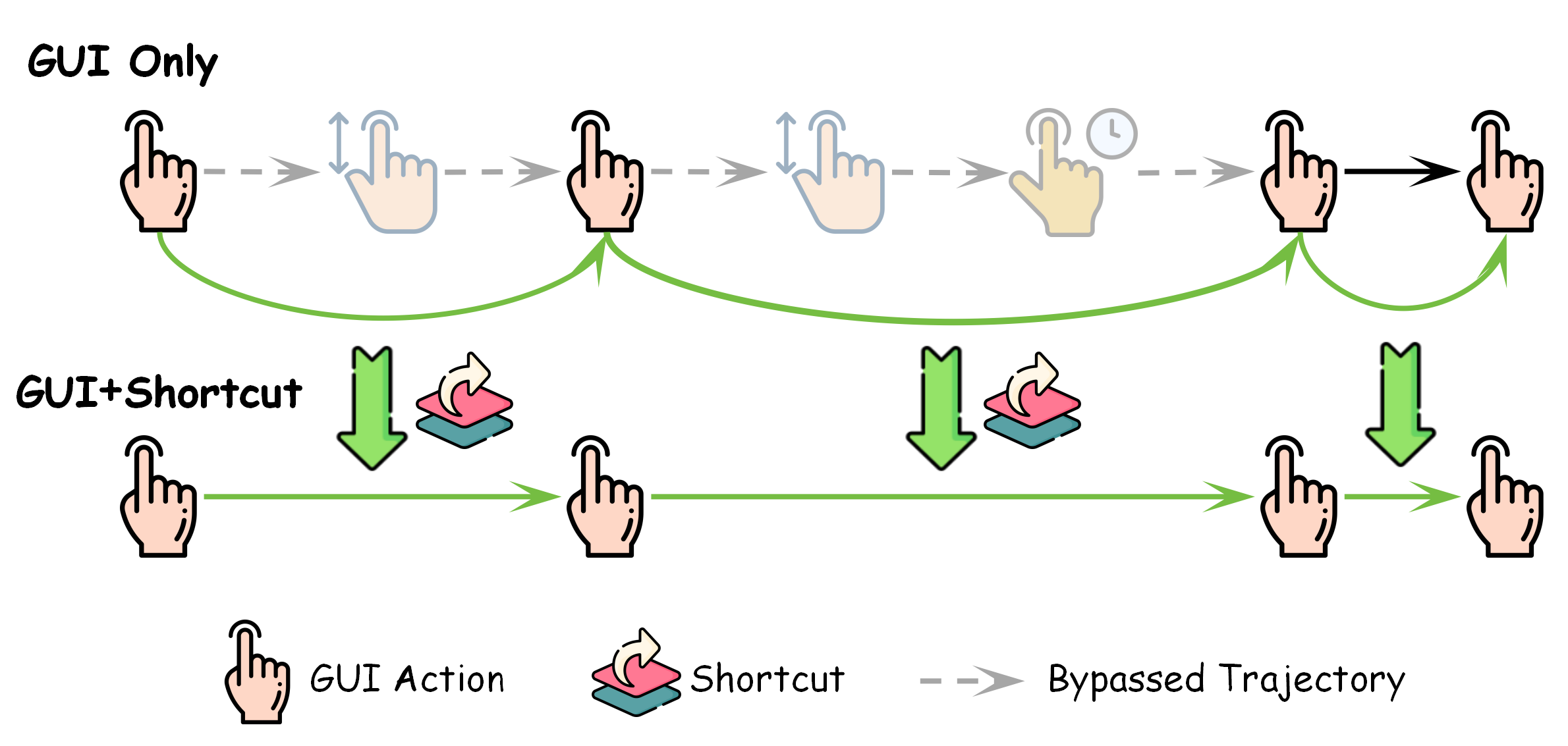}
  \caption{\textbf{Workflow of GUI Only vs. GUI-Shortcut Hybrid Agent.} Shortcuts improve agent execution efficiency by bypassing GUI operations.}
  \label{fig:teaser}
\end{figure}

As GUI agents are increasingly applied to complex tasks, enhancing their efficiency has become a central research focus. In this context, hybrid paradigms have emerged as a promising solution, demonstrating effectiveness across a wide range of platforms~\cite{zhang2025api,guo2026e3tirenhancedexperienceexploitation}. This approach combines the speed and reliability of ``shortcuts'' such as Application Programming Interface (API) calls, deep links, and Robotic Process Automation (RPA) scripts~\cite{agostinelli2022reactive}, with the flexibility of GUI operations. As Fig.~\ref{fig:teaser} illustrates, a hybrid agent bypasses multi-step GUI operations by invoking a single shortcut, drastically reducing operational complexity and time.

However, despite these advancements, a framework for systematically evaluating and benchmarking hybrid agents is still lacking~\cite{zhang2025api,wang2025mcpflowfacilitatingllmagents,huang2025critictool}. This gap leaves the full potential of GUI-Shortcut agents far from being thoroughly explored and assessed. Mobile platforms provide a controlled and reproducible setting, making them a natural starting point for GUI-shortcut evaluation.

Therefore, we present \textbf{MAS-Bench}, \textbf{the first benchmark explicitly designed to evaluate GUI-shortcut hybrid mobile agents.} The tasks within MAS-Bench are solvable entirely through GUI interactions, but shortcuts can substantially improve execution efficiency. MAS-Bench evaluates whether agents can reliably choose between GUI actions and shortcuts under dynamic mobile environments. Our experiments show that MAS-GLM-4.5V achieves the best success rate of 68.3\%. In a matched comparison, MAS-MobileAgent improves over its GUI-only counterpart from 35.2\% to 63.3\% SR, yielding a 79.8\% relative gain and reducing the successful-task step ratio by 38.9\%.

Furthermore, \textbf{we introduce a novel framework to evaluate agents' capacity to generate new shortcuts from interaction.} This framework integrates agent-generated shortcuts into a standard baseline agent and measures subsequent task performance. Our findings reveal a performance gap: while our predefined shortcuts prove highly reliable (100\% success rate), agent-generated shortcuts lag in robustness and efficiency. Dynamic shortcuts demonstrate significant potential. Despite their task completion rate of 38\% (baseline 43\%), they offer the highest efficiency, highlighting them as a promising direction for future research into robust shortcut generation. 

Our contributions are threefold:

\begin{itemize}
  \item We introduce \textbf{MAS-Bench}, the first benchmark for systematically evaluating GUI-shortcut hybrid mobile agents. It comprises 139 complex tasks spanning 11 real-world applications, supported by a knowledge base of 88 predefined shortcuts and 9 distinct evaluation metrics.
  \item We establish extensive baselines across agentic workflows, general-purpose models, and specialized GUI models, demonstrating that GUI-shortcut hybrid operation substantially improves success rate and efficiency while exposing shortcut misuse.
  \item We propose the first framework to evaluate an agent's ability to generate shortcuts from interaction trajectories. Using predefined structural shortcuts as a reference upper-bound baseline, our experiments show that current generated shortcuts still lag behind predefined ones in efficiency and robustness, highlighting a key direction for future work.

\end{itemize}

\section{Related Work}

\paragraph{Mobile Task Automation.}

Mobile task automation evolves from methods based on predefined scripts to more intelligent and adaptive agents driven by LLMs~\cite{li2025mtr,wang2025fedmabench}. Traditional approaches, such as API calls, deep links, and Robotic Process Automation (RPA) scripts, offer direct execution paths but suffer from significant limitations, including invalidation from app updates, and a sensitivity to UI changes that hinders their ability to adapt~\cite{kennedy2011use,xiao2025ui}.

The rapid development of LLM-based GUI agents significantly overcomes traditional automation challenges by simulating human interaction through visual or multimodal inputs~\cite{cheng2024seeclick}. However, complete reliance on LLMs for fine-grained GUI interaction presents new problems: step-by-step action for routine tasks leads to inefficiency and increased costs; in complex operations, LLM hallucinations or misinterpretations can cause cumulative errors, affecting task success rates and reliability~\cite{wen2024autodroidv2}. Zhang et al.~\cite{zhang2025api} provide a detailed analysis of API and GUI agents; the former gain attention for their stability and programming integration capabilities, while the latter possesses strong adaptability. These considerations regarding the efficiency and robustness of GUI agents lead to an exploration of how to combine LLM with efficient shortcut methods, thereby promoting the rise of research in GUI-Shortcut agents.

\paragraph{GUI-Shortcut Hybrid Operations.}

Researchers explore GUI-Shortcut methods to enhance agent capabilities and address the persistent challenges in task completion rate, costs, and operational efficiency for LLM-based GUI agents. For instance, UFO2~\cite{zhang2025ufo2}, a desktop AgentOS, demonstrates robust task execution through its multi-agent architecture and a unified GUI-API action layer that deeply integrates with operating system and application features. Inspired by such work, AppAgentX~\cite{jiang2025appagentx} improves efficiency by evolving high-level actions into shortcuts based on task execution history. MobileAgent-E~\cite{wang2025mobile} uses a self-evolving module to learn and store general-purpose \textit{Tips} and reusable \textit{Shortcuts} from past experiences, continuously improving performance and efficiency over time. While these efforts focused on the mobile domain are promising, they also highlight the urgent need for a systematic method to evaluate the effectiveness of these emerging hybrid operations on mobile devices.

\paragraph{Mobile GUI Agent Benchmark.}

As agents evolve from GUI-based interaction to hybrid GUI-Shortcut operations, they place new demands on benchmarking. Among existing benchmarks, MobileAgentBench~\cite{wang2024mobileagentbench} focuses on the foundational GUI navigation and task completion capabilities of LLM-powered mobile agents. AndroidWorld and SPA-Bench~\cite{rawles2024androidworld,chen2024spa} benchmark mobile agents on diverse smartphone tasks, spanning dynamic environments, task difficulty. While these benchmarks advance research in various dimensions, there remains a need for a benchmark that can comprehensively evaluate the ability of agents to intelligently discover, decide upon, and execute diverse shortcuts within GUI interactions and also measure the overall task effectiveness of this GUI-Shortcut hybrid operation~\cite{zhang2025ufo2}.
\begin{table}[!t]
\centering
\resizebox{0.99\linewidth}{!}{%
\begin{tabular}{lc>{\centering\arraybackslash}p{0.45cm}>{\centering\arraybackslash}p{0.45cm}>{\centering\arraybackslash}p{0.45cm}>{\centering\arraybackslash}p{0.45cm}}
\toprule
\textbf{Benchmark} & \textbf{\# Inst.} & \textbf{DS} & \textbf{DL} & \textbf{EC} & \textbf{SG} \\
\midrule
AndroidArena~\cite{xing2024AndroidArena}
& 221 & \crossmy & \crossmy & \crossmy  & \crossmy \\
AndroidWorld~\cite{rawles2024androidworld}
& 116 & \crossmy & \checkmarkmy  & \crossmy  & \crossmy \\
LlamaTouch~\cite{zhang2024llamatouch}
& 495 & \crossmy & \checkmarkmy & \crossmy  & \crossmy \\
AndroidLab~\cite{xu2024androidlab}
& 138 & \crossmy & \crossmy & \checkmarkmy  & \crossmy \\
SPA-Bench~\cite{chen2024spa}
& 340 & \crossmy & \checkmarkmy & \checkmarkmy  & \crossmy \\
\midrule
\rowcolor[HTML]{EFEFEF}
\textbf{\system(Ours)} & 139 & \checkmarkmy & \checkmarkmy  & \checkmarkmy & \checkmarkmy \\
\bottomrule
\end{tabular}
}
\caption{\textbf{Comparison of \system\ and other smartphone agent benchmarks with dynamic environments.} Column definitions: \# Inst. (number of instructions), \# DS (support diverse shortcuts), \# DL (difficulty level), \# EC (efficiency\&cost metrics), \# SG (evaluate shortcut generation).}
\label{tab:benchmark-comparison}
% \vspace{-0.2cm}
\end{table}

\section{MAS-Bench}\label{sec:masbench}

\subsection{Online Evaluation Environment}

MAS-Bench is built on a dynamic Android platform to evaluate mobile agent performance on complex tasks comprehensively. Unlike static benchmarks, which are often limited to fixed datasets, our platform enables the real-time assessment of agents' capacity to select and utilize shortcuts intelligently. To guarantee reproducibility, we employ a snapshot-based reset mechanism that restores the environment to a consistent initial state after each task, and utilize dedicated test accounts to isolate real-world side effects (detailed in Appendix~\ref{app:environment_control}).

\subsection{GUI-Shortcut Hybrid Action Space}

To complete tasks, agents in MAS-Bench operate within a hybrid action space that combines conventional GUI interactions with diverse programmatic shortcuts. This space comprises two action types:

\paragraph{GUI Action.} GUI actions simulate direct human interaction with an application's interface. This action space consists of basic operations such as click, swipe, and type, enabling the agent to navigate and interact with UI elements. Detailed definitions for each action are provided in Appendix~\ref{app:action_space}. GUI actions are essential for dynamic scenarios with unavailable predefined shortcuts.

\paragraph{Shortcut Action.} Shortcut actions can trigger specific functions or navigate to designated pages, bypassing multi-step GUI actions. In MAS-Bench, shortcuts are categorized into two types: \textbf{Predefined shortcuts} and \textbf{Agent-generated shortcuts}.

\begin{figure}[t]
    \centering
    \includegraphics[width=0.9\columnwidth]{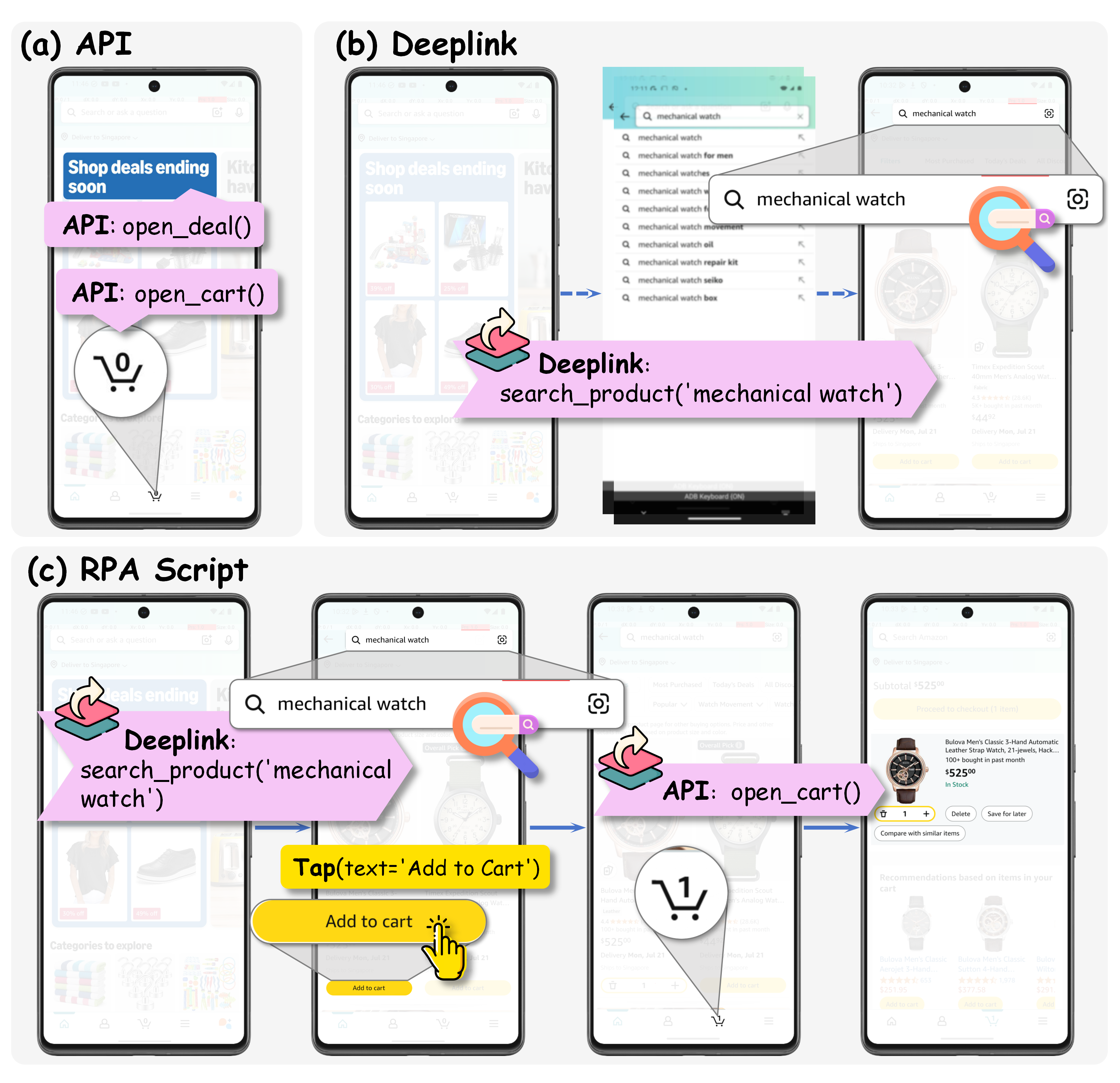}
    \caption{\textbf{Functional Comparison of APIs, Deep-Links, and RPA Scripts.} The figure uses the Amazon app as an example: (a) the \textit{open\_cart()} API directly opens the shopping cart; (b) the \textit{search\_product()} deep-link directly performs a product search; and (c) an RPA script combines APIs, deep links, and GUI operations to automate a complete workflow.}
    \label{fig:api_deeplink_rpa}
\end{figure}

As illustrated in Fig.~\ref{fig:api_deeplink_rpa}, the predefined shortcuts in \system primarily consist of APIs, deep-links, and RPA scripts.
\begin{itemize}
\item \textbf{API}: An intent-based programmatic shortcut that invokes application functionality without GUI interaction, as shown in Fig.~\ref{fig:api_deeplink_rpa}(a).
\item \textbf{Deep Link}: A specialized URI that targets a specific page or function within an application. As shown in Fig.~\ref{fig:api_deeplink_rpa}(b), this enables direct navigation, bypassing multi-step GUI actions.
\item \textbf{RPA Script}: An Automation script designed to handle a specific, highly repetitive subtask. Unlike a single action, an RPA script encapsulates a complex workflow of GUI actions, API calls, or deep-links, consolidating a multi-step process into a single, efficient shortcut action, as shown in Fig.~\ref{fig:api_deeplink_rpa}(c).
\end{itemize}

The agent-generated shortcuts are created dynamically by the agent itself. They are typically formed by identifying and abstracting repetitive subtasks from the agent's execution history into new, executable routines. This capability allows the agent to learn from experience and progressively optimize its performance across future tasks by creating customized shortcuts.

\begin{figure*}[ht]
    \centering
    \includegraphics[width=\textwidth]{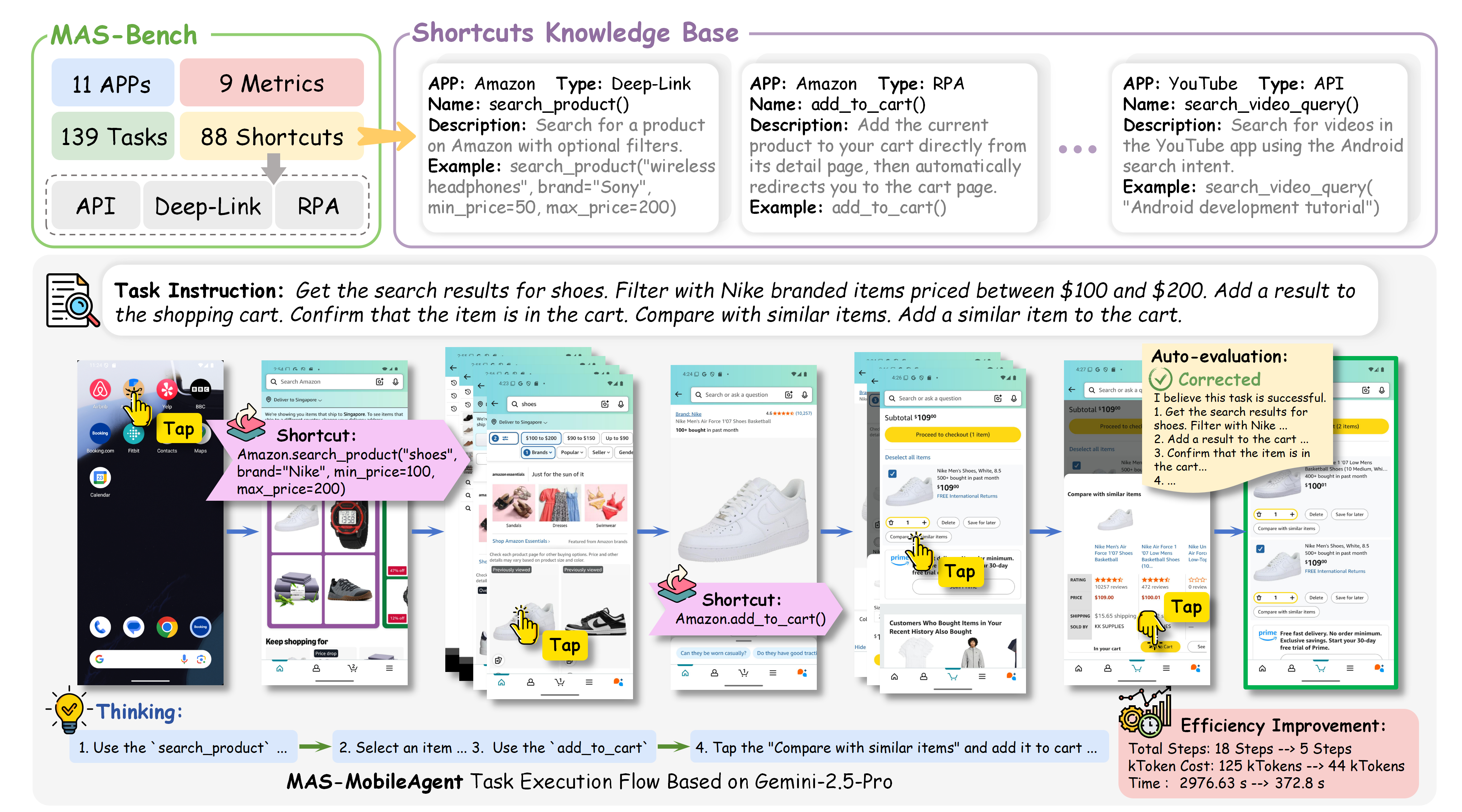}
    \caption{\textbf{The pipeline of MAS-Bench}. The GUI-Shortcut agent first filters products using the \textit{search\_product} shortcut, selects an item via GUI operations, and then adds it to the cart using the \textit{add\_to\_cart} shortcut. The entire process is monitored by an automated evaluation module, which outputs metrics such as success rate and efficiency.}
    \label{fig:pipeline}
\end{figure*}

\subsection{Shortcut Knowledge Base}
\label{sec:shortcut_knowledge_base}

\paragraph{Predefined Shortcuts Knowledge Base.} To validate the feasibility of hybrid GUI-shortcut operations on mobile devices, we construct a predefined knowledge base of commonly used mobile shortcuts. Based on 11 popular apps, it includes 11 APIs, 70 Deep Links, and 7 custom RPA scripts, for a total of 88 predefined shortcuts. The distribution of shortcuts across applications varies based on app complexity and functionality, ranging from 2 shortcuts for Google Calendar to 21 for Fitbit (see Appendix Table~\ref{table:app-shortcut}). To ensure real-world relevance, APIs were identified from official documentation and static analysis of application packages. These APIs target intent-accessible endpoints and are executed through standard Android IPC/ADB commands without requiring root access, debug signatures, or emulator system modifications. At the same time, Deep Links were identified by analyzing each application's declared URI schemes and entry points. The custom RPA scripts were designed to encapsulate common, high-repetition sub-tasks, testing an agent's ability to leverage complex, pre-built automation routines. More details on the collection and design methodology are available in the Appendix~\ref{app:predefined_shortcut_collection}.

\paragraph{Agent-Generated Shortcuts.} Besides predefined shortcuts, MAS-Bench supports evaluating agent-generated shortcuts to test an agent's learning and abstraction capabilities. In order to demonstrate the effectiveness of MAS-Bench in evaluating agents' capabilities of generating shortcuts, we have incorporated several contrasting shortcut generation methods for evaluation. We include Macro-level action trajectory replay (abbreviated as action replay) shortcuts that record and replay entire task or sub-task execution trajectories. We also incorporate dynamic shortcuts that require the agent to perform real-time grounding based on execution history, as well as shortcuts generated by the MobileAgent-E~\cite{wang2025mobile}. This diverse set of generation strategies allows for a comprehensive assessment of an agent's ability to autonomously create efficient workflows.

\subsection{Tasks Design for GUI-Shortcut Hybrid Agent}

MAS-Bench comprises 139 complex tasks derived from real-world scenarios across 11 Android applications, each featuring one or more invocable predefined shortcuts. These tasks reflect authentic user needs, with an average of 9.27 steps for single-app and 17.66 for cross-app workflows based on human operation. Tasks are categorized into three difficulty levels (detailed in Appendix~\ref{app:task_design}), with category-level examples summarized in Table~\ref{tab:task_category_examples}.

Specifically, tasks in MAS-Bench maintain high real-world relevance. They are designed by considering user needs from daily life, such as searching for products in a shopping app or navigating to a location. The tasks typically involve multiple steps and sub-goals, and usable shortcuts exist within our established knowledge base for each task. However, our shortcut knowledge base for these complex tasks is intentionally incomplete. It offers only basic shortcuts, which are insufficient to complete an entire task independently. This design compels agents to synthesize solutions by combining shortcut invocations with multi-step GUI operations, thus providing a more realistic test of their planning and reasoning abilities.

\begin{figure}[t]
    \centering
    \includegraphics[width=0.9\columnwidth]{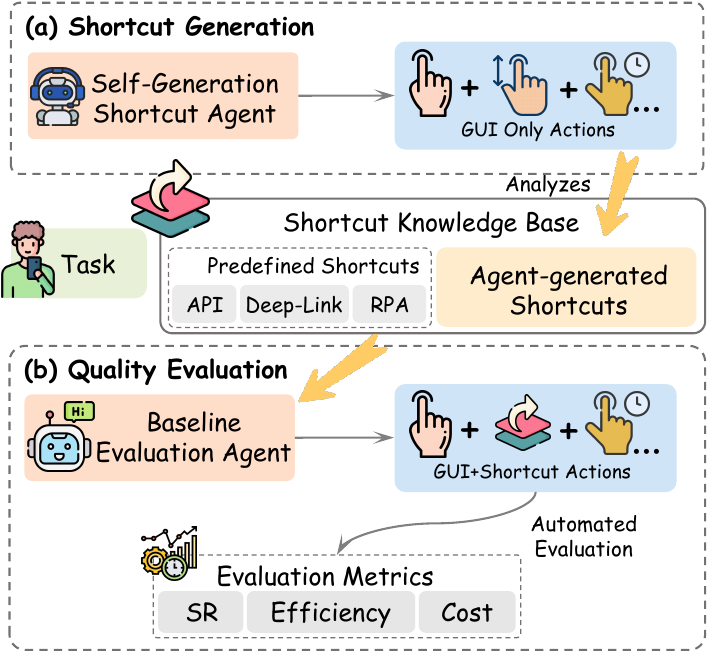}
    \caption{\textbf{Evaluation Workflow for Agents' Shortcut Generation Capability.} The process consists of two stages: (a) Shortcut Generation Stage, where the agent creates its shortcut knowledge base; and (b) Quality Evaluation Stage. The agent-generated shortcut knowledge base is imported into a baseline agent for performance testing. Its quality is then measured by comparing this performance against the GUI-only baseline agent and the baseline agent with predefined shortcuts.}
    \label{fig:generation_evaluation}
\end{figure}

To test the agent's learning and generation capabilities, we introduce scenarios with repetitive sub-tasks. In these scenarios, the knowledge base intentionally lacks predefined shortcuts for these recurrent operations. This design aims to test whether an agent, after completing a sub-task via GUI operations for the first time, can identify the repetitive pattern and then autonomously generate an efficient shortcut for subsequent use. Notably, some tasks in MAS-Bench are designed as incremental variations, where a task builds upon the previous one by adding additional requirements. This incremental design serves two purposes: it creates realistic scenarios where agents must handle evolving user needs, and it provides an ideal testbed for evaluating shortcut generation by presenting repeated sub-task patterns that agents should learn to abstract into reusable shortcuts.

\subsection{Systematic Hybrid Agent Evaluation}

\paragraph{GUI-Shortcut Hybrid Operation Evaluation.} MAS-Bench is designed to evaluate an agent's ability to autonomously discover and utilize shortcuts to improve task success, efficiency, and cost-effectiveness. This evaluation mandates that an agent possess sophisticated recognition and decision-making capabilities. Agents must not only identify available shortcuts based on the current task and environment but also critically assess their applicability and efficiency. Furthermore, the agent must choose between GUI operations and shortcuts to reduce steps and time while avoiding irrelevant shortcut calls.

\paragraph{Shortcut Generation Evaluation.} 
We expect GUI agents not only to rely on predefined shortcuts within the knowledge base but also to possess the capability for shortcut generation. The quality of agent-generated shortcuts serves as a direct indicator of their generative capability. However, a systematic framework to develop and evaluate the effectiveness of these generation methods is currently lacking.
Therefore, MAS-Bench encourages and evaluates the agent's ability to generate new shortcuts autonomously. As shown in Fig.~\ref{fig:generation_evaluation}, we design the following evaluation method. First, a self-generation shortcut agent explores the MAS-Bench environment and creates its shortcut knowledge base. To ensure fairness and eliminate interference from the intrinsic abilities of the agent under test, we use a unified baseline GUI agent as the evaluated agent and import these shortcut knowledge bases into the baseline agent for task execution. The baseline agent's performance on the tasks directly reflects the quality of the imported knowledge base, thus enabling an evaluation of each agent's shortcut generation capabilities.

\subsection{Evaluation Metrics}
\label{sec:evaluation_metrics}

We employ 9 comprehensive metrics to evaluate agent performance across three dimensions: (1) \textit{Success}: Success Rate (SR) measures task completion; (2) \textit{Efficiency}: Mean Steps (MS), Mean Step Ratio (MSR), Mean Step Ratio on Successful tasks (MSRS), and Mean Execution Time (MET) assess operational efficiency; (3) \textit{Cost and Resource Utilization}: Mean Token Cost (MToC), Mean Shortcut Call Count (MSC), Shortcut Success Rate (SSR), and Shortcut-to-GUI Ratio (S2GR) evaluate resource consumption and operational strategy. Table~\ref{table:metrics_summary} in Appendix~\ref{app:metrics_summary} provides detailed definitions, units, and improvement directions for all metrics.

\subsection{Evaluation Methodology}
\label{sec:eval_method}

We introduce \textbf{MAS-Bench-Eval} as a scalable and reproducible evaluation protocol for MAS-Bench. Illustrated in Fig.~\ref{fig:evaluation_pipeline}, this automated pipeline adopts a two-stage \textit{Describe-and-Judge} framework:

\textbf{Stage 1: Action Semantics Extraction.} An efficient MLLM processes the agent's trajectory step-by-step, generating textual captions that describe the executed actions and corresponding UI state transitions. This stage transforms the raw visual stream into a structured semantic history.

\textbf{Stage 2: Success Determination.} A highly capable MLLM acts as the judge. It reasons over the task instruction, the semantic history derived from Stage 1, and the terminal screenshot to issue a binary success verdict.

This decoupled architecture effectively optimizes the trade-off between inference efficiency and evaluation capabilities. To validate its reliability, we compare MAS-Bench-Eval against 278 human-annotated trajectories and obtain high agreement with human judgments, with an F1 score of 96.3\% and precision of 98.1\%. We provide the full analysis in Appendix~\ref{app:eval_reliability}.

\section{MAS: Shortcut-Augmented Hybrid Mobile Agents}\label{sec:baseline}

To evaluate the efficacy of the GUI-Shortcut hybrid operation, we introduce MAS-MobileAgent, which serves as a reference for GUI-shortcut Hybrid Agents. Built upon MobileAgent-V2~\cite{wang2024mobileagentv2}, which relies on visual perception for UI understanding, MAS-MobileAgent augments its decision-making process by retrieving relevant shortcuts from the knowledge base and integrating them with visual inputs.

To further assess the generalization of our shortcut injection mechanism, we introduce MAS-T3A. Derived from T3A~\cite{rawles2024androidworld}, this agent operates on a distinct modality, utilizing structured UI trees rather than visual screenshots. By extending the hybrid operation to this text-based framework, we aim to verify its transferability to non-visual perceptual foundations.

Finally, to validate the effectiveness of hybrid operation within the \textit{Agent-as-a-Model} paradigm, we apply the same shortcut injection strategy to multiple model families, including Qwen3-VL~\cite{bai2025qwen3}, GLM-4.5V~\cite{hong2025glm}, ScaleCUA~\cite{liu2025scalecua}, and MAI-UI~\cite{zhou2025mai}. Relevant shortcuts are incorporated directly into the system prompt without modifying the model weights. Collectively, these diverse implementations enable a rigorous and comprehensive assessment of our framework's universality across mainstream agent architectures.

\section{Experiments}\label{sec:experiment}
\begin{table*}[!ht]
\centering
\resizebox{0.99\textwidth}{!}{%
\begin{tabular}{lccccccccc}
\toprule
\multirow{3}{*}{Agent} & \multicolumn{2}{c}{Input} & \multirow{3}{*}{\begin{tabular}[c]{@{}c@{}}SR$\uparrow$\end{tabular}} & \multicolumn{3}{c}{Efficiency} & \multicolumn{2}{c}{Cost} & \multirow{3}{*}{\begin{tabular}[c]{@{}c@{}}S2GR$\uparrow$\end{tabular}} \\
\cmidrule(lr){2-3} \cmidrule(lr){5-7} \cmidrule(lr){8-9}
    & \begin{tabular}[c]{@{}c@{}}SS\end{tabular} & \begin{tabular}[c]{@{}c@{}}VH\end{tabular} & & \begin{tabular}[c]{@{}c@{}}MS$\downarrow$\end{tabular} & \begin{tabular}[c]{@{}c@{}}MSRS$\downarrow$\end{tabular} & \begin{tabular}[c]{@{}c@{}}MET$\downarrow$\end{tabular} & \begin{tabular}[c]{@{}c@{}}MToC$\downarrow$\end{tabular} & \begin{tabular}[c]{@{}c@{}}MSC$\uparrow$\end{tabular} &  \\
\midrule
\textit{Human}  & \checkmarkmy &  & - & \textit{12.11} & \textit{1.000} & - & - & - & - \\
\midrule
\rowcolor[HTML]{E8E8E8}
\multicolumn{10}{c}{\textit{Agentic Workflow (Gemini-2.5-Pro)}} \\
\midrule
M3A~\cite{rawles2024androidworld} & \checkmarkmy & \checkmarkmy & 0.503 & 16.655 & 1.131 & 266.505 & 200.509 & 0 & 0 \\
MobileAgent-E~\cite{wang2025mobile} & \checkmarkmy &  & 0.259 & \textbf{4.808} & 0.857 & 462.636 & \textbf{87.819} & 1.011 & 0.114 \\
\cmidrule(lr){1-10}
T3A~\cite{rawles2024androidworld} & & \checkmarkmy  & 0.453 & 15.755 & 1.067 & \underline{178.005} & 440.919 & 0 & 0 \\
\rowcolor[HTML]{E6F2FF}
\quad\textbf{+ MAS-T3A (Ours)} & & \checkmarkmy & \underline{0.554}{\small\textcolor{blue}{$_{\uparrow22.3\%}$}} & \underline{12.768}{\small\textcolor{blue}{$_{\uparrow19.0\%}$}} & \underline{0.823}{\small\textcolor{blue}{$_{\uparrow22.9\%}$}} & \textbf{148.505}{\small\textcolor{blue}{$_{\uparrow16.6\%}$}} & 341.402{\small\textcolor{blue}{$_{\uparrow23\%}$}} & \underline{1.438} & \underline{0.140} \\
\cmidrule(lr){1-10}
MobileAgentV2~\cite{wang2024mobileagentv2} & \checkmarkmy &  & 0.352 & 19.252 & 1.122 & 1364.979 & 156.441 & 0 & 0 \\
\rowcolor[HTML]{E6F2FF}
\quad\textbf{+ MAS-MobileAgent (Ours)} & \checkmarkmy &  & \textbf{0.633}{\small\textcolor{blue}{$_{\uparrow79.8\%}$}} & 12.928{\small\textcolor{blue}{$_{\uparrow32.8\%}$}} & \textbf{0.686}{\small\textcolor{blue}{$_{\uparrow38.9\%}$}} & 951.918{\small\textcolor{blue}{$_{\uparrow30.3\%}$}} & \underline{130.980}{\small\textcolor{blue}{$_{\uparrow16\%}$}} & \textbf{1.948} & \textbf{0.341} \\
\midrule
\rowcolor[HTML]{E8E8E8}
\multicolumn{10}{c}{\textit{General-Purpose Models}} \\
\midrule
Qwen2.5-VL-3B~\cite{bai2025qwen2} & \checkmarkmy &  & 0.036 & 21.626 & 1.341 & \textbf{120.975} & - & 0 & 0 \\
Qwen2.5-VL-7B~\cite{bai2025qwen2} & \checkmarkmy &  & 0.022 & 21.835 & 1.498 & 168.832 & - & 0 & 0 \\
\cmidrule(lr){1-10}
Qwen3-VL-4B~\cite{bai2025qwen3} & \checkmarkmy &  & 0.228 & 17.547 & 1.273 & 179.208 & - & 0 & 0 \\
\rowcolor[HTML]{E6F2FF}
\quad\textbf{+ MAS-Qwen3-VL-4B (Ours)} & \checkmarkmy &  & 0.237{\small\textcolor{blue}{$_{\uparrow3.9\%}$}} & 20.403{\small\textcolor{red}{$_{-16.3\%}$}} & 0.971{\small\textcolor{blue}{$_{\uparrow23.7\%}$}} & 188.078{\small\textcolor{red}{$_{-4.9\%}$}} & - & 2.461 & 0.170 \\
\cmidrule(lr){1-10}
Qwen3-VL-8B~\cite{bai2025qwen3} & \checkmarkmy &  & 0.259 & 16.576 & 1.183 & 179.377 & - & 0 & 0 \\
\rowcolor[HTML]{E6F2FF}
\quad\textbf{+ MAS-Qwen3-VL-8B (Ours)} & \checkmarkmy &  & 0.425{\small\textcolor{blue}{$_{\uparrow64.1\%}$}} & 14.921{\small\textcolor{blue}{$_{\uparrow10.0\%}$}} & 0.868{\small\textcolor{blue}{$_{\uparrow26.6\%}$}} & 155.977{\small\textcolor{blue}{$_{\uparrow13.0\%}$}} & - & 1.309 & 0.110 \\
\cmidrule(lr){1-10}
Qwen3-VL-32B~\cite{bai2025qwen3} & \checkmarkmy &  & 0.338 & 17.834 & 1.207 & 394.102 & - & 0 & 0 \\
\rowcolor[HTML]{E6F2FF}
\quad\textbf{+ MAS-Qwen3-VL-32B (Ours)} & \checkmarkmy &  & 0.446{\small\textcolor{blue}{$_{\uparrow32.0\%}$}} & 15.913{\small\textcolor{blue}{$_{\uparrow10.8\%}$}} & \underline{0.848}{\small\textcolor{blue}{$_{\uparrow29.7\%}$}} & 358.282{\small\textcolor{blue}{$_{\uparrow9.1\%}$}} & - & \textbf{3.216} & \textbf{0.276} \\
\cmidrule(lr){1-10}
Qwen3-VL-235B~\cite{bai2025qwen3} & \checkmarkmy &  & 0.417 & 16.604 & 1.111 & 185.201 & - & 0 & 0 \\
\rowcolor[HTML]{E6F2FF}
\quad\textbf{+ MAS-Qwen3-VL-235B (Ours)} & \checkmarkmy &  & 0.525{\small\textcolor{blue}{$_{\uparrow25.9\%}$}} & \underline{14.424}{\small\textcolor{blue}{$_{\uparrow13.1\%}$}} & \textbf{0.785}{\small\textcolor{blue}{$_{\uparrow29.3\%}$}} & \underline{152.652}{\small\textcolor{blue}{$_{\uparrow17.6\%}$}} & - & \underline{2.784} & \underline{0.233} \\
\cmidrule(lr){1-10}
GLM-4.5V~\cite{hong2025glm} & \checkmarkmy &  & \underline{0.526} & 17.237 & 1.194 & 281.928 & - & 0 & 0 \\
\rowcolor[HTML]{E6F2FF}
\quad\textbf{+ MAS-GLM-4.5V (Ours)} & \checkmarkmy &  & \textbf{0.683}{\small\textcolor{blue}{$_{\uparrow29.8\%}$}} & \textbf{14.050}{\small\textcolor{blue}{$_{\uparrow18.5\%}$}} & 0.900{\small\textcolor{blue}{$_{\uparrow24.6\%}$}} & 238.579{\small\textcolor{blue}{$_{\uparrow15.4\%}$}} & - & 1.051 & 0.097 \\
\midrule
\rowcolor[HTML]{E8E8E8}
\multicolumn{10}{c}{\textit{Specialized GUI Models}} \\
\midrule
UI-TARS-1.5-7B~\cite{ui-tars-15} & \checkmarkmy &  & 0.287 & 19.209 & \underline{1.191} & 188.143 & - & 0 & 0 \\
GUI-Owl-7B~\cite{mobileagentv3} & \checkmarkmy &  & 0.295 & \textbf{15.568} & 1.239 & 168.148 & - & 0 & 0 \\
\cmidrule(lr){1-10}
ScaleCUA-7B~\cite{liu2025scalecua} & \checkmarkmy &  & 0.108 & 18.194 & 1.202 & \textbf{123.018} & - & 0 & 0 \\
\rowcolor[HTML]{E6F2FF}
\quad\textbf{+ MAS-ScaleCUA-7B (Ours)} & \checkmarkmy &  & 0.115{\small\textcolor{blue}{$_{\uparrow6.5\%}$}} & 19.446{\small\textcolor{red}{$_{-6.9\%}$}} & 1.358{\small\textcolor{red}{$_{-13.0\%}$}} & 141.535{\small\textcolor{red}{$_{-15.1\%}$}} & - & \underline{0.007} & 0.000 \\
\cmidrule(lr){1-10}
ScaleCUA-32B~\cite{liu2025scalecua} & \checkmarkmy &  & 0.231 & 19.209 & 1.286 & 141.654 & - & 0 & 0 \\
\rowcolor[HTML]{E6F2FF}
\quad\textbf{+ MAS-ScaleCUA-32B (Ours)} & \checkmarkmy &  & 0.216{\small\textcolor{red}{$_{-6.5\%}$}} & 18.935{\small\textcolor{blue}{$_{\uparrow1.4\%}$}} & 1.260{\small\textcolor{blue}{$_{\uparrow2.0\%}$}} & \underline{140.126}{\small\textcolor{blue}{$_{\uparrow1.1\%}$}} & - & 0.000 & 0.000 \\
\cmidrule(lr){1-10}
MAI-UI-8B~\cite{zhou2025mai} & \checkmarkmy &  & \underline{0.489} & 19.237 & 1.247 & 180.678 & - & 0 & 0 \\
\rowcolor[HTML]{E6F2FF}
\quad\textbf{+ MAS-MAI-UI-8B (Ours)} & \checkmarkmy &  & \textbf{0.583}{\small\textcolor{blue}{$_{\uparrow19.2\%}$}} & \underline{18.065}{\small\textcolor{blue}{$_{\uparrow6.1\%}$}} & \textbf{1.179}{\small\textcolor{blue}{$_{\uparrow5.5\%}$}} & 169.463{\small\textcolor{blue}{$_{\uparrow6.2\%}$}} & - & \textbf{0.273} & \textbf{0.018} \\
\bottomrule
\end{tabular}
}
\caption{\textbf{Overall performance comparison on MAS-Bench with predefined shortcuts knowledge base (139 tasks).} Results are weighted averages based on task distribution (92 single-app, 47 cross-app tasks). Bold and underlined values denote the best and second-best results within each block, respectively. Methods marked with ``\textbf{+}'' are shortcut-augmented variants of their corresponding baselines. SS: Screenshot; VH: View Hierarchy; MS: Mean Steps; MSRS: Mean Step Ratio on Successful tasks; MET: Mean Execution Time (seconds); MToC: Mean Token Cost (in thousands); MSC: Mean Shortcut Call count; S2GR: Shortcut-to-GUI action Ratio. Detailed results in Appendix~\ref{app:detailed_results}.}
\label{tab:results-masbench}
% \vspace{-0.2cm}
\end{table*}

We conduct the experiments in the online, dynamic environment of MAS-Bench and evaluate the performance of GUI-shortcut hybrid mobile GUI agents. We also employ an evaluation framework to measure the efficiency of agent-generated and predefined shortcuts.

\subsection{Experiment Setup}

We evaluate GUI agents on the 139 tasks in MAS-Bench, including three categories: 1) GUI-only agents, covering Agentic Workflows (e.g., T3A, M3A~\cite{rawles2024androidworld}, and MobileAgentV2~\cite{wang2024mobileagentv2}) and Agent-as-a-Model methods, including General-Purpose Models (e.g., Qwen3-VL~\cite{bai2025qwen3} and GLM-4.5V~\cite{hong2025glm}) and Specialized GUI Models (e.g., UI-TARS-1.5~\cite{ui-tars-15}, GUI-Owl~\cite{mobileagentv3}, ScaleCUA~\cite{liu2025scalecua}, and MAI-UI~\cite{zhou2025mai}); 2) self-generated shortcut frameworks (MobileAgent-E~\cite{wang2025mobile}); 3) shortcut-augmented Hybrid Agents (GUI-shortcut agents), including MAS-T3A, MAS-MobileAgent, and MAS variants of Qwen3-VL, GLM-4.5V, ScaleCUA, and MAI-UI. For agentic workflows, we use Gemini-2.5-Pro as the base model. To examine the impact of predefined shortcuts across models with different capabilities, we also evaluate Gemini-2.0-Flash. All evaluations follow the metrics defined in Sec.~\ref{sec:evaluation_metrics} and Appendix~\ref{app:environment_setting}.

\subsection{Evaluation Protocols}

\paragraph{Predefined Shortcut Evaluation.}
To evaluate GUI-shortcut hybrid operation, we augment each compatible GUI agent with the predefined shortcut knowledge base and compare it with its GUI-only counterpart under the same task set, environment, and evaluation metrics.

\paragraph{Shortcut Invocation Robustness.}
To evaluate whether agents reliably select shortcuts, we annotate ground-truth shortcuts for each task and compare them with those invoked by agents. We report Mean Task Shortcut Recall (MTSR), Mean Shortcut Selection Precision (MSSP), MF1, and redundancy rate. To simulate long-tail cases where no suitable shortcut, we construct an interference setting on 23 cross-app tasks by removing ground-truth shortcuts and exposing task-irrelevant distractors. We report the changes in SR and MS, together with Mean Irrelevant Shortcut Calls (MISC). See Appendix~\ref{app:shortcut_invocation_robustness} for details.

\paragraph{Shortcut Generation Evaluation.}
To assess the quality of shortcut knowledge bases generated by different strategies, we conduct a two-stage evaluation on a randomly selected test subset comprising 50\% of the tasks in MAS-Bench.
In the \textbf{Shortcut Generation Stage}, we construct several types of agent-generated shortcuts to form knowledge bases using two methods. One leverages MobileAgent-E to generate shortcuts ($S_{\text{MobileAgent-E}}$). The other, based on execution trajectories of M3A, generates three types of shortcuts: task-level action replays ($S_{\text{Replay-Task}}$), subtask-level action replays ($S_{\text{Replay-Subtask}}$), and dynamic shortcuts ($S_{\text{Dynamic}}$) that require the model to perform grounding. Details of the shortcut knowledge base are provided in Sec.~\ref{sec:shortcut_knowledge_base}.
In the \textbf{Quality Evaluation Stage}, each knowledge base is mapped to a standardized action space and integrated into the T3A baseline agent. This method treats the baseline agent's performance as a reflection of the imported knowledge base's quality, enabling an unbiased, executor-controlled evaluation of the generation strategies.

\subsection{Effectiveness of GUI-Shortcut Operation}

Based on our experimental results in Table~\ref{tab:results-masbench}, we identify four key findings. Additional single-app and cross-app breakdowns for representative agents are provided in Appendix~\ref{app:detailed_results}.

\paragraph{Finding 1: Integrating a predefined shortcut knowledge base significantly enhances agent performance, validating the effectiveness of the hybrid GUI-shortcut operation.} Experiments show that using the predefined knowledge base of 88 shortcuts significantly improves task success rate and efficiency across multiple agent families. Compared to MobileAgentV2, MAS-MobileAgent improves the success rate (SR) from 35.2\% to 63.3\%, an 80\% improvement, while reducing the token cost from 156.441k to 130.980k. The gain also holds for Agent-as-a-Model approaches: MAS-GLM-4.5V achieves the best overall SR of 68.3\%, and MAS-Qwen3-VL-8B improves SR from 25.9\% to 42.5\%. Furthermore, the Mean Step Ratio on Successful tasks (MSRS) serves as a robust efficiency metric, particularly for mitigating the impact of premature termination in long-horizon tasks. MAS-MobileAgent achieves an MSRS of 0.686, which is significantly lower than the baseline version's 1.122, indicating that its execution path is closer to the optimal solution. This also highlights the efficiency gap between current GUI-only agents and the human pure-GUI step baseline: strong GUI-only baselines remain above MSRS $=1.0$, while hybrid agents can reduce MSRS below 1.0 by bypassing GUI paths with shortcuts. The SR and MS results for Different-level Tasks can be found in the Appendix~\ref{app:different_level_tasks}.

\paragraph{Finding 2: The effectiveness of predefined shortcuts is framework-agnostic, benefiting agents with different input modalities.} Experiments show that both agentic workflows based on structured UI trees (MAS-T3A) and raw screenshots (MAS-MobileAgent), as well as Agent-as-a-Model variants (e.g., MAS-Qwen3-VL and MAS-GLM-4.5V), benefit from the same predefined shortcut knowledge base. This independence stems from the fact that shortcut execution is decoupled from the agent's perception modality: shortcuts directly send commands to the operating system or application, reducing the need to parse the UI layout in real time. The results further show that General-Purpose Models adapt better to the hybrid action space, whereas specialized GUI models such as ScaleCUA do not automatically benefit without reliable shortcut selection.

\begin{figure}[t]
  \centering
  \includegraphics[width=\columnwidth]{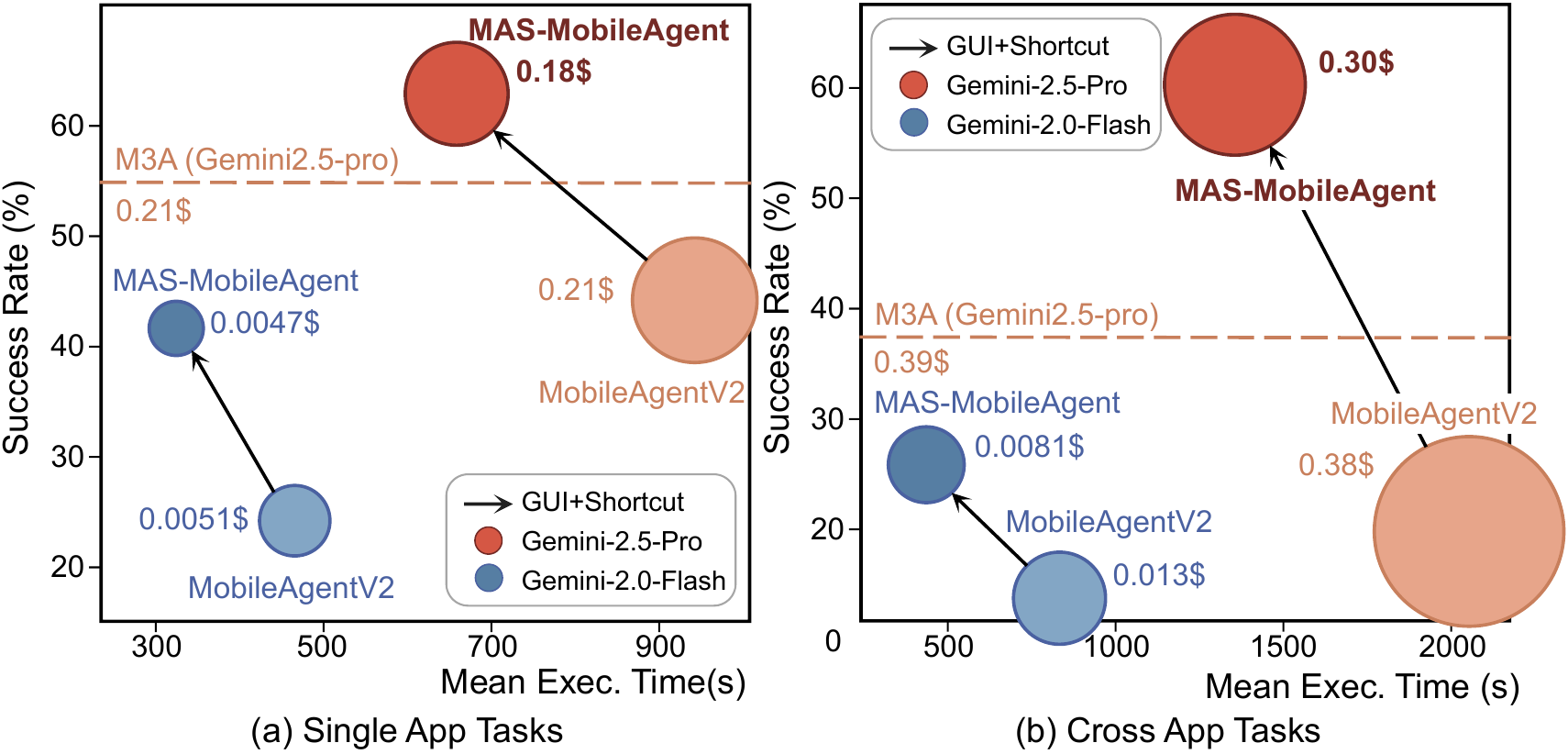}
  \caption{\textbf{Performance comparison of MAS-MobileAgent with and without shortcuts.} The base models are Gemini-2.5-Pro and Gemini-2.0-Flash. Data points show the relationship between SR and MET for single-app and cross-app tasks, with circle size representing mean cost. Results demonstrate that shortcuts benefit both models, with more significant improvements for the weaker Gemini-2.0-Flash.}
  \label{fig:pro_flash_diff}
\end{figure}

\paragraph{Finding 3: Shortcut gains depend on base model capability and shortcut-use reliability.} The Qwen3-VL series shows that shortcut gains are not monotonic with model scale: Qwen3-VL-8B achieves the largest relative SR gain (+64.1\%), while Qwen3-VL-4B suffers an efficiency drop (+16.3\% MS). This indicates that hybrid operation requires sufficient reasoning ability to select and invoke shortcuts reliably. At the same time, shortcuts can substantially help weaker but capable agents bypass failure-prone GUI steps; for example, augmenting Gemini-2.0-Flash boosts its cross-app SR from 0\% to 23.4\% (Appendix~\ref{app:diff_pro_flash}).

\paragraph{Finding 4: MAS-Bench exposes and penalizes shortcut misuse.} Shortcut selection quality directly affects performance, as detailed in Appendix~\ref{app:shortcut_invocation_robustness}. GLM-4.5V maintains low redundancy (4.8\%) and reduces MS by 18.5\%, whereas Qwen3-VL-4B has higher redundancy (33.5\%) and increases MS by 16.3\% (Table~\ref{tab:shortcut_selection_quality}). In the no-relevant-shortcut interference setting, Qwen3-VL-4B suffers a 66.7\% relative SR drop under irrelevant shortcuts, while GLM-4.5V drops only 7.7\% (Table~\ref{tab:shortcut_interference}). These results show that MAS-Bench can evaluate and quantify agent robustness under noisy shortcut pools, not only performance when relevant shortcuts are available.

\subsection{Results of Shortcut Generation}

Our experiments include a baseline T3A framework without shortcuts and five distinct shortcut variants: Predefined ($S_{\text{Predefined}}$), Replay-Task ($S_{\text{Replay-Task}}$), Replay-Subtask ($S_{\text{Replay-Subtask}}$), Dynamic ($S_{\text{Dynamic}}$), and MobileAgent-E ($S_{\text{MobileAgent-E}}$). The predefined shortcut knowledge base serves as a reference upper bound for high-quality structural shortcuts, rather than a target that current automatic methods are expected to surpass. This setup allows us to compare shortcut generation paradigms under the same executor and quantify how close generated shortcuts are to reliable human-designed shortcuts.

In our experiments, Predefined shortcuts achieve the best performance, improving the success rate (SR) by 9\% over the baseline while maintaining a perfect shortcut success rate (SSR) of 100\%. Notably, agents using Predefined shortcuts reduce average execution steps by 25\% and decrease total execution time by approximately 16\%. Despite having the second-highest shortcut call count per task (1.45), Predefined shortcuts maintain excellent robustness, executing successfully across diverse task configurations. The increased success rate and reduced execution time validate that well-designed shortcuts will enhance GUI task efficiency.

In contrast, Replay-Task shortcuts exhibit poor robustness with only a 10\% success rate, indicating high susceptibility to environmental variations. This fragility negatively impacts task completion rates and execution efficiency, underscoring that while effective shortcuts boost GUI performance, poorly designed ones are counterproductive.

Results reveal substantial room for improvement in model-generated shortcuts. Across variants, only Predefined and MobileAgent-E shortcuts improve model accuracy, with only Predefined shortcuts simultaneously enhancing execution speed and success rate. These results indicate current limitations in shortcut generation capabilities within our framework. Future work should focus on developing more efficient, robust shortcuts with higher utilization rates that effectively reduce execution steps.

\begin{table}[!t]
\centering
\resizebox{0.99\linewidth}{!}{%
\begin{tabular}{lccccc}
\toprule
\multirow{2}{*}{Shortcut} & \multirow{2}{*}{\begin{tabular}[c]{@{}c@{}}SR$\uparrow$\end{tabular}} & \multirow{2}{*}{\begin{tabular}[c]{@{}c@{}}SSR$\uparrow$\end{tabular}} & \multirow{2}{*}{\begin{tabular}[c]{@{}c@{}}MSRS$\downarrow$\end{tabular}} & \multirow{2}{*}{\begin{tabular}[c]{@{}c@{}}MSC$\uparrow$\end{tabular}}  & \multirow{2}{*}{\begin{tabular}[c]{@{}c@{}}MET$\downarrow$\end{tabular}} \\
& & & & & \\
\midrule
\textit{Human} & - & - & \textit{1.00} & - & - \\
\midrule
\rowcolor[HTML]{EFEFEF}
\textit{Baseline} & 0.43 & - & 0.96 & - & 188.93 \\
\midrule
$S_{\text{Predefined}}$ & \textbf{0.52} & \textbf{1.00} & \textbf{0.71} & 1.45  & \textbf{152.15}\\
$S_{\text{Replay-Task}}$ & 0.34 & 0.10 & 0.91 & \textbf{3.04} & 244.61 \\
$S_{\text{Replay-Subtask}}$ & 0.43 & 0.73 & 1.13 & 1.22 & 236.67\\
$S_{\text{Dynamic}}$ & 0.38 & 0.75 & 0.82 & 0.91  & 216.24  \\
$S_{\text{MobileAgent-E}}$ & 0.49 & 0.71 & 1.00 & 1.01 & 224.87 \\
\bottomrule
\end{tabular}
}
\caption{\textbf{The results of different shortcut generation methods.} Column definitions: \# SR (success rate), \# MSRS (Mean Step Ratio on Successful tasks), \# MSC (Mean Shortcut Call Count), \# SSR (Shortcut Success Rate), \# MET (Mean Execution Time).}
\label{tab:generation_result}
% \vspace{-0.2cm}
\end{table}

\section{Conclusion} 
\label{sec:conclusion}

In this paper, we introduce MAS-Bench, the first unified and comprehensive benchmark designed to evaluate the effectiveness of GUI-shortcut hybrid mobile agents. Furthermore, MAS-Bench provides a framework to assess the quality and validity of shortcuts that are autonomously generated by the agent. Our experiments demonstrate that the GUI-shortcut hybrid operation significantly enhances both the success rate and efficiency of task execution. Moreover, we also validate that our benchmark effectively measures the quality and robustness of the agent-generated shortcuts. We hope MAS-Bench will support future work on efficient GUI-shortcut hybrid mobile agents, including agent self-improvement through iterative shortcut accumulation, validation, and reuse.

\section{Limitations}
\label{sec:limitations}
While our predefined shortcuts demonstrate robustness across application versions (Appendix~\ref{app:predefined_shortcut_collection}), agents should maintain the capability to fall back to GUI operations when shortcuts occasionally fail due to environmental variations. Additionally, our evaluation on standardized emulators ensures reproducibility and fair comparison; future work will validate these findings on real-world devices with diverse configurations. Another direction is to integrate dynamic shortcut discovery into MAS-Bench, e.g., by mining AndroidManifest files, static analysis results, official documentation, or web resources to construct structural shortcuts automatically. Finally, although MAS-Bench-Eval shows strong agreement with human judgments, future versions can combine rule-based state checks with LLM-as-a-Judge to further improve evaluation stability. This work highlights the importance of efficiency in mobile GUI agents and calls for further research into GUI-shortcut hybrid approaches.

% Bibliography entries for the entire Anthology, followed by custom entries
%\bibliography{custom,anthology-overleaf-1,anthology-overleaf-2}

% Custom bibliography entries only
\bibliography{custom}

@article{huang2025critictool,
  title={CRITICTOOL: Evaluating Self-Critique Capabilities of Large Language Models in Tool-Calling Error Scenarios},
  author={Huang, Shiting and Fang, Zhen and Chen, Zehui and Yuan, Siyu and Ye, Junjie and Zeng, Yu and Chen, Lin and Mao, Qi and Zhao, Feng},
  journal={arXiv preprint arXiv:2506.13977},
  year={2025}
}

@article{li2026no,
  title={No More Stale Feedback: Co-Evolving Critics for Open-World Agent Learning},
  author={Li, Zhicong and Jiang, Lingjie and Hu, Yulan and Zeng, Xingchen and Li, Yixia and Zhang, Xiangwen and Chen, Guanhua and Pan, Zheng and Li, Xin and Liu, Yong},
  journal={arXiv preprint arXiv:2601.06794},
  year={2026}
}

@article{jiang2025think,
  title={Think only when you need with large hybrid-reasoning models},
  author={Jiang, Lingjie and Wu, Xun and Huang, Shaohan and Dong, Qingxiu and Chi, Zewen and Dong, Li and Zhang, Xingxing and Lv, Tengchao and Cui, Lei and Wei, Furu},
  journal={arXiv preprint arXiv:2505.14631},
  year={2025}
}

@article{xiao2025ui,
  title={Ui-genie: A self-improving approach for iteratively boosting mllm-based mobile gui agents},
  author={Xiao, Han and Wang, Guozhi and Chai, Yuxiang and Lu, Zimu and Lin, Weifeng and He, Hao and Fan, Lue and Bian, Liuyang and Hu, Rui and Liu, Liang and others},
  journal={arXiv preprint arXiv:2505.21496},
  year={2025}
}

@article{xiao2026ui,
  title={UI-Mem: Self-Evolving Experience Memory for Online Reinforcement Learning in Mobile GUI Agents},
  author={Xiao, Han and Wang, Guozhi and Wang, Hao and Liu, Shilong and Chai, Yuxiang and Pan, Yue and Zhou, Yufeng and Chen, Xiaoxin and Wen, Yafei and Li, Hongsheng},
  journal={arXiv preprint arXiv:2602.05832},
  year={2026}
}

@misc{guo2026e3tirenhancedexperienceexploitation,
      title={E3-TIR: Enhanced Experience Exploitation for Tool-Integrated Reasoning},
      author={Weiyang Guo and Zesheng Shi and Liye Zhao and Jiayuan Ma and Zeen Zhu and Junxian He and Min Zhang and Jing Li},
      year={2026},
      eprint={2604.09455},
      archivePrefix={arXiv},
      primaryClass={cs.AI},
      url={https://arxiv.org/abs/2604.09455},
}

@article{li2025mtr,
  title={Mtr-bench: A comprehensive benchmark for multi-turn reasoning evaluation},
  author={Li, Xiaoyuan and Bao, Keqin and Ma, Yubo and Li, Moxin and Wang, Wenjie and Men, Rui and Zhang, Yichang and Feng, Fuli and Liu, Dayiheng and Lin, Junyang},
  journal={arXiv preprint arXiv:2505.17123},
  year={2025}
}

@article{wang2025think,
  title={Think-While-Generating: On-the-Fly Reasoning for Personalized Long-Form Generation},
  author={Wang, Chengbing and Zhang, Yang and Wang, Wenjie and Zhao, Xiaoyan and Feng, Fuli and He, Xiangnan and Chua, Tat-Seng},
  journal={arXiv preprint arXiv:2512.06690},
  year={2025}
}

@inproceedings{wang2025fedmabench,
  title={FedMABench: Benchmarking Mobile GUI Agents on Decentralized Heterogeneous User Data},
  author={Wang, Wenhao and Yu, Zijie and Ye, Rui and Zhang, Jianqing and Liu, Guangyi and Liu, Liang and Chen, Siheng and Wang, Yanfeng},
  booktitle={Proceedings of the 2025 Conference on Empirical Methods in Natural Language Processing},
  pages={26398--26419},
  year={2025}
}

@inproceedings{huang2025cogddn,
  title={Cogddn: A cognitive demand-driven navigation with decision optimization and dual-process thinking},
  author={Huang, Yuehao and Liu, Liang and Lei, Shuangming and Ma, Yukai and Su, Hao and Mei, Jianbiao and Zhao, Pengxiang and Gu, Yaqing and Liu, Yong and Lv, Jiajun},
  booktitle={Proceedings of the 33rd ACM International Conference on Multimedia},
  pages={5237--5246},
  year={2025}
}

@article{lu2025uis1,
  title={Ui-s1: Advancing gui automation via semi-online reinforcement learning},
  author={Lu, Zhengxi and Ye, Jiabo and Tang, Fei and Shen, Yongliang and Xu, Haiyang and Zheng, Ziwei and Lu, Weiming and Yan, Ming and Huang, Fei and Xiao, Jun and others},
  journal={arXiv preprint arXiv:2509.11543},
  year={2025}
}

@misc{ui-tars-15,
  title = {UI-TARS-1.5},
  author = {ByteDance Seed},
  howpublished = {\url{https://seed-tars.com/1.5}},
  year = {2025},
}

@article{bai2025qwen2,
  title={Qwen2. 5-vl technical report},
  author={Bai, Shuai and Chen, Keqin and Liu, Xuejing and Wang, Jialin and Ge, Wenbin and Song, Sibo and Dang, Kai and Wang, Peng and Wang, Shijie and Tang, Jun and others},
  journal={arXiv preprint arXiv:2502.13923},
  year={2025}
}

@article{bai2025qwen3,
  title={Qwen3-vl technical report},
  author={Bai, Shuai and Cai, Yuxuan and Chen, Ruizhe and Chen, Keqin and Chen, Xionghui and Cheng, Zesen and Deng, Lianghao and Ding, Wei and Gao, Chang and Ge, Chunjiang and others},
  journal={arXiv preprint arXiv:2511.21631},
  year={2025}
}

@article{hong2025glm,
  title={Glm-4.5 v and glm-4.1 v-thinking: Towards versatile multimodal reasoning with scalable reinforcement learning},
  author={Hong, Wenyi and Yu, Wenmeng and Gu, Xiaotao and Wang, Guo and Gan, Guobing and Tang, Haomiao and Cheng, Jiale and Qi, Ji and Ji, Junhui and Pan, Lihang and others},
  journal={arXiv preprint arXiv:2507.01006},
  year={2025}
}

@article{liu2025scalecua,
  title={Scalecua: Scaling open-source computer use agents with cross-platform data},
  author={Liu, Zhaoyang and Xie, JingJing and Ding, Zichen and Li, Zehao and Yang, Bowen and Wu, Zhenyu and Wang, Xuehui and Sun, Qiushi and Liu, Shi and Wang, Weiyun and others},
  journal={arXiv preprint arXiv:2509.15221},
  year={2025}
}

@article{zhou2025mai,
  title={MAI-UI Technical Report: Real-World Centric Foundation GUI Agents},
  author={Zhou, Hanzhang and Zhang, Xu and Tong, Panrong and Zhang, Jianan and Chen, Liangyu and Kong, Quyu and Cai, Chenglin and Liu, Chen and Wang, Yue and Zhou, Jingren and others},
  journal={arXiv preprint arXiv:2512.22047},
  year={2025}
}

@article{mobileagentv3,
  title={Mobile-agent-v3: Fundamental agents for gui automation},
  author={Ye, Jiabo and Zhang, Xi and Xu, Haiyang and Liu, Haowei and Wang, Junyang and Zhu, Zhaoqing and Zheng, Ziwei and Gao, Feiyu and Cao, Junjie and Lu, Zhengxi and others},
  journal={arXiv preprint arXiv:2508.15144},
  year={2025}
}

@misc{wang2025mcpflowfacilitatingllmagents,
      title={MCP-Flow: Facilitating LLM Agents to Master Real-World, Diverse and Scaling MCP Tools},
      author={Wenhao Wang and Peizhi Niu and Zhao Xu and Zhaoyu Chen and Jian Du and Yaxin Du and Xianghe Pang and Keduan Huang and Yanfeng Wang and Qiang Yan and Siheng Chen},
      year={2025},
      eprint={2510.24284},
      archivePrefix={arXiv},
      primaryClass={cs.AI},
      url={https://arxiv.org/abs/2510.24284},
}

@article{jiang2025appagentx,
  title={AppAgentX: Evolving GUI Agents as Proficient Smartphone Users},
  author={Jiang, Wenjia and Zhuang, Yangyang and Song, Chenxi and Yang, Xu and Zhou, Joey Tianyi and Zhang, Chi},
  journal={arXiv preprint arXiv:2503.02268},
  year={2025}
}

@inproceedings{lu2026ui,
  title={Ui-r1: Enhancing efficient action prediction of gui agents by reinforcement learning},
  author={Lu, Zhengxi and Chai, Yuxiang and Guo, Yaxuan and Yin, Xi and Liu, Liang and Wang, Hao and Xiao, Han and Ren, Shuai and Zhao, Pengxiang and Liu, Guangyi and others},
  booktitle={Proceedings of the AAAI Conference on Artificial Intelligence},
  volume={40},
  number={21},
  pages={17608--17616},
  year={2026}
}

@article{liu2025learnact,
  title={LearnAct: Few-Shot Mobile GUI Agent with a Unified Demonstration Benchmark},
  author={Liu, Guangyi and Zhao, Pengxiang and Liu, Liang and Chen, Zhiming and Chai, Yuxiang and Ren, Shuai and Wang, Hao and He, Shibo and Meng, Wenchao},
  journal={arXiv preprint arXiv:2504.13805},
  year={2025}
}

@article{agostinelli2022reactive,
  title={Reactive synthesis of software robots in RPA from user interface logs},
  author={Agostinelli, Simone and Lupia, Marco and Marrella, Andrea and Mecella, Massimo},
  journal={Computers in Industry},
  volume={142},
  pages={103721},
  year={2022},
  publisher={Elsevier}
}

@article{kennedy2011use,
  title={Use of cognitive shortcuts in landline and cell phone surveys},
  author={Kennedy, Courtney and Everett, Stephen E},
  journal={Public Opinion Quarterly},
  volume={75},
  number={2},
  pages={336--348},
  year={2011},
  publisher={Oxford University Press}
}

@article{xu2024androidlab,
  title={AndroidLab: Training and Systematic Benchmarking of Android Autonomous Agents},
  author={Xu, Yifan and Liu, Xiao and Sun, Xueqiao and Cheng, Siyi and Yu, Hao and Lai, Hanyu and Zhang, Shudan and Zhang, Dan and Tang, Jie and Dong, Yuxiao},
  journal={arXiv preprint arXiv:2410.24024},
  year={2024}
}

@article{cheng2024seeclick,
  title={Seeclick: Harnessing gui grounding for advanced visual gui agents},
  author={Cheng, Kanzhi and Sun, Qiushi and Chu, Yougang and Xu, Fangzhi and Li, Yantao and Zhang, Jianbing and Wu, Zhiyong},
  journal={arXiv preprint arXiv:2401.10935},
  year={2024}
}

@inproceedings{xing2024AndroidArena,
  title={Understanding the weakness of large language model agents within a complex android environment},
  author={Xing, Mingzhe and Zhang, Rongkai and Xue, Hui and Chen, Qi and Yang, Fan and Xiao, Zhen},
  booktitle={Proceedings of the 30th ACM SIGKDD Conference on Knowledge Discovery and Data Mining},
  pages={6061--6072},
  year={2024}
}

@article{zhang2024llamatouch,
  title={LlamaTouch: A Faithful and Scalable Testbed for Mobile UI Automation Task Evaluation},
  author={Zhang, Li and Wang, Shihe and Jia, Xianqing and Zheng, Zhihan and Yan, Yunhe and Gao, Longxi and Li, Yuanchun and Xu, Mengwei},
  journal={arXiv preprint arXiv:2404.16054},
  year={2024}
}

@article{wang2024mobileagentv2,
  title={Mobile-Agent-v2: Mobile Device Operation Assistant with Effective Navigation via Multi-Agent Collaboration},
  author={Wang, Junyang and Xu, Haiyang and Jia, Haitao and Zhang, Xi and Yan, Ming and Shen, Weizhou and Zhang, Ji and Huang, Fei and Sang, Jitao},
  journal={arXiv preprint arXiv:2406.01014},
  year={2024}
}

@article{wang2024mobileagentbench,
  title={MobileAgentBench: An Efficient and User-Friendly Benchmark for Mobile LLM Agents},
  author={Wang, Luyuan and Deng, Yongyu and Zha, Yiwei and Mao, Guodong and Wang, Qinmin and Min, Tianchen and Chen, Wei and Chen, Shoufa},
  journal={arXiv preprint arXiv:2406.08184},
  year={2024}
}

@article{rawles2024androidworld,
  title={AndroidWorld: A dynamic benchmarking environment for autonomous agents},
  author={Rawles, Christopher and Clinckemaillie, Sarah and Chang, Yifan and Waltz, Jonathan and Lau, Gabrielle and Fair, Marybeth and Li, Alice and Bishop, William and Li, Wei and Campbell-Ajala, Folawiyo and others},
  journal={arXiv preprint arXiv:2405.14573},
  year={2024}
}

@article{wang2025mobile,
  title={Mobile-Agent-E: Self-Evolving Mobile Assistant for Complex Tasks},
  author={Wang, Zhenhailong and Xu, Haiyang and Wang, Junyang and Zhang, Xi and Yan, Ming and Zhang, Ji and Huang, Fei and Ji, Heng},
  journal={arXiv preprint arXiv:2501.11733},
  year={2025}
}

@article{wen2024autodroidv2,
  title={AutoDroid-V2: Boosting SLM-based GUI Agents via Code Generation},
  author={Wen, Hao and Tian, Shizuo and Pavlov, Borislav and Du, Wenjie and Li, Yixuan and Chang, Ge and Zhao, Shanhui and Liu, Jiacheng and Liu, Yunxin and Zhang, Ya-Qin and others},
  journal={arXiv preprint arXiv:2412.18116},
  year={2024}
}

@misc{openai2024gpt4technicalreport,
      title={GPT-4 Technical Report},
      author={OpenAI},
      year={2024},
      eprint={2303.08774},
      archivePrefix={arXiv},
      primaryClass={cs.CL},
      url={https://arxiv.org/abs/2303.08774},
}

@article{liu2025llm,
  title={Llm-powered gui agents in phone automation: Surveying progress and prospects},
  author={Liu, Guangyi and Zhao, Pengxiang and Liu, Liang and Guo, Yaxuan and Xiao, Han and Lin, Weifeng and Chai, Yuxiang and Han, Yue and Ren, Shuai and Wang, Hao and others},
  journal={arXiv preprint arXiv:2504.19838},
  year={2025}
}

@article{zhang2025api,
  title={API Agents vs. GUI Agents: Divergence and Convergence},
  author={Zhang, Chaoyun and He, Shilin and Li, Liqun and Qin, Si and Kang, Yu and Lin, Qingwei and Zhang, Dongmei},
  journal={arXiv preprint arXiv:2503.11069},
  year={2025}
}

@article{zhang2025ufo2,
  title={Ufo2: The desktop agentos},
  author={Zhang, Chaoyun and Huang, He and Ni, Chiming and Mu, Jian and Qin, Si and He, Shilin and Wang, Lu and Yang, Fangkai and Zhao, Pu and Du, Chao and others},
  journal={arXiv preprint arXiv:2504.14603},
  year={2025}
}

@inproceedings{chen2024spa,
  title={Spa-bench: A comprehensive benchmark for smartphone agent evaluation},
  author={Chen, Jingxuan and Yuen, Derek and Xie, Bin and Yang, Yuhao and Chen, Gongwei and Wu, Zhihao and Yixing, Li and Zhou, Xurui and Liu, Weiwen and Wang, Shuai and others},
  booktitle={NeurIPS 2024 Workshop on Open-World Agents},
  year={2024}
}

\newpage
\appendix
\clearpage
\section*{Appendix}

\section{MAS-Bench Environment}
\label{app:environment_setting}
\subsection{Observation Space}

MAS-Bench evaluates mobile agents in a highly standardized and controllable environment using an Android Virtual Device (AVD). Specifically, the environment is based on a Pixel 7 emulator running Android 15 (API Level 35). Its screen resolution is set to 1080 x 2400 pixels, and the screen density is 420 dpi. The emulator has 24GB of RAM and 24GB of storage. The environment support parallel execution of multi-threaded evaluations. All model deployments and evaluations are conducted on 4 NVIDIA L40S GPUs 48GB.

\subsection{Environment Control and Reproducibility}
\label{app:environment_control}

MAS-Bench's dynamic evaluation approach moves beyond simply verifying task outcomes, focusing on the agent's decision-making process in live scenarios. To ensure real-world relevance, the benchmark incorporates a suite of widely used Android applications covering diverse daily-life scenarios. Ensuring consistent and reproducible evaluation environments is critical for fair agent comparison, particularly when working with dynamic online applications. MAS-Bench addresses this challenge through a snapshot-based rapid recovery mechanism combined with carefully controlled test accounts.

\paragraph{Snapshot-Based Rapid Recovery.} 
We leverage Android emulator snapshot functionality to guarantee absolute environment consistency across evaluations. Prior to experimentation, we construct a pre-configured Android Virtual Device image, termed \textbf{MAS-Bench-AVD}, which includes all 11 benchmark applications with necessary permissions granted and dedicated test accounts logged in. This clean baseline state is saved as a snapshot. We maintain a single system-level snapshot rather than task- or app-specific snapshots, so this design introduces no additional per-task storage overhead beyond the base AVD image.

\paragraph{Online Content Control.} 
For applications involving dynamic online content (e.g., Amazon, YouTube, Gmail), we employ pre-registered anonymous dedicated test accounts to maintain a relatively stable recommendation baseline. These accounts are isolated from real user data. For tasks involving email communication, we utilize anonymous test accounts configured with automatic reply functionality, ensuring that all email-based tasks interact only within our controlled test environment. This design prevents interference with real-world users while maintaining realistic task scenarios.

\paragraph{Parallel Execution and Scalability.} 
To support large-scale evaluation, MAS-Bench implements ADB port-based isolation for managing multiple emulator instances concurrently. Each emulator instance operates independently with its own snapshot rollback mechanism, enabling parallel task execution without cross-contamination. This isolation strategy is equally effective for shortcut actions, ensuring that they can be executed concurrently without interference.

\paragraph{Real-World Side Effect Mitigation.} 
All benchmark tasks execute within our controlled test environment using dedicated test accounts. Actions such as adding an item to cart, or sending emails affect only these isolated test accounts and never impact real users or production systems. This design ensures ethical evaluation practices while maintaining task authenticity.

\subsection{Details of Action Space}
\label{app:action_space}

\begin{table}[h]
    \centering
    \resizebox{0.99\linewidth}{!}{%
    \begin{tabular}{>{\raggedright\arraybackslash}p{2.5cm} >{\raggedright\arraybackslash}p{5.2cm}}
    \toprule
    \textbf{Action} & \textbf{Definition} \\
    \midrule
    Tap ($x$, $y$) & Tap at coordinates ($x$, $y$). \\
    \midrule
    Type (text) & Enter text content into an input field. \\
    \midrule
    Swipe ($x_1$, $y_1$, $x_2$, $y_2$) & Swipe from ($x_1$, $y_1$) to ($x_2$, $y_2$). \\
    \midrule
    Home & Go to the home screen. \\
    \midrule
    Back & Go back to the previous app screen. \\
    \midrule
    Stop & Complete the current task. \\
    \bottomrule
    \end{tabular}
    }
    \caption{Examples of the basic action space in MAS-Bench.}
    \label{table:action-space}
\end{table} 

The MAS-Bench environment facilitates agent-device interaction through a low-level interface based on the Android Debug Bridge (ADB). Crucially, the specific action space available to an agent during evaluation ultimately depends on its architecture and capabilities. Table~\ref{table:action-space} shows examples of the basic action set.

\subsection{Details of Predefined Shortcuts Knowledge Base}

\paragraph{Details of Predefined Shortcuts.}
\label{app:predefined_shortcut_collection}

Our shortcut knowledge base was created to provide agents with an authentic yet challenging environment. We collect and design shortcuts for 11 selected applications, ensuring their relevance to real-world usage patterns. Importantly, we focus on collecting shortcuts that represent \textbf{commonly used functionalities} of these applications, which tend to be stable across app versions.

\begin{itemize}
    \item \textbf{API Collection}: The APIs in our knowledge base are intent-accessible endpoints rather than private or non-exported APIs. We identify documented endpoints from official Android documentation and use static analysis of application packages to locate additional endpoints exposed through standard Android IPC. We retain only APIs that can be invoked by standard ADB commands, such as \texttt{adb shell am start}, with Intent actions, data URIs, and extras. They do not require root access, debug signatures, hidden permissions, or emulator system modifications.
    \item \textbf{Deep-Links Collection}: We collect deep-links by analyzing each application's \texttt{AndroidManifest.xml} file. This file typically declares the URL schemes and paths the application can handle, providing direct mappings to specific pages or functions.
    \item \textbf{RPA Script Design}: RPA scripts are designed for common, highly repetitive sub-tasks (e.g., adding an item with specific options to a shopping cart). These scripts are implemented using UI tree-based element identification, which provides better robustness against UI layout changes compared to coordinate-based approaches. The scripts can be executed as a single action, serving as a high-level shortcut. This design enables us to test an agent's ability to utilize complex, pre-built automation routines.
\end{itemize}

\paragraph{Stability of Shortcuts.} 
\label{app:shortcut_stability}
The stability of shortcuts across application versions is a critical consideration for real-world deployment. Our collection strategy deliberately focuses on core, commonly used functionalities that applications typically maintain across updates. \textbf{API and Deep-Link shortcuts} exhibit high stability, as they interface with core application functions that rarely change across versions. These shortcuts are based on documented or intent-accessible endpoints and declared URI schemes, which developers typically maintain for backward compatibility. \textbf{RPA scripts} in our knowledge base are implemented using UI tree-based element identification rather than fixed coordinates. This approach leverages the structured view hierarchy, which remains more stable across application updates compared to pixel-based coordinate systems, resulting in enhanced robustness against UI layout changes.

To validate the robustness of our shortcut knowledge base, we have tested all shortcuts across at least five different versions of each application, spanning updates released over six months. Our validation confirms that all shortcuts in our knowledge base remain functional across these versions without modification, demonstrating their practical viability for real-world agent deployment. 

\paragraph{Details of Agent-Generated Shortcuts.} 
\label{app:agent_generated_shortcuts}
We incorporate several methods to create agent-generated shortcuts. The primary purpose for including this variety of shortcut types is to validate the effectiveness of the MAS-Bench framework itself in evaluating an agent's shortcut generation capabilities. By creating test cases with diverse characteristics, we can verify our benchmark's ability to assess different shortcut generation strategies.

\begin{itemize}
    \item \textbf{Macro-level action trajectory replay (action replay) Shortcuts}: This category of shortcuts is constructed by recording and replaying an agent's historical action trajectories. They execute a fixed sequence of low-level actions, such as clicks and swipes at specific coordinates, to automate a previously completed workflow. This method is implemented at two distinct granularities:
    \begin{itemize}
        \item \textbf{Task-Level Replay}: Captures and replays the entire action sequence of a full task, from start to finish. This results in a single, high-level macro that is highly specific to one complete workflow.
        \item \textbf{Subtask-Level Replay}: Identifies and abstracts recurrent, multi-step operations within a larger task. This approach generates shorter, more modular shortcuts that automate common sub-workflows.
    \end{itemize}
    \item \textbf{Dynamic Shortcuts}: Unlike static macro-replays, these shortcuts are adaptive and robust to UI changes. Instead of replaying hardcoded coordinates, they utilize real-time grounding to semantically identify and interact with target UI elements based on the current screen context. This allows them to function correctly even if the position or layout of UI elements changes.
\end{itemize}
Ultimately, this diverse suite of shortcut types validates that MAS-Bench provides a comprehensive framework for evaluating agents' shortcut generation capabilities.

\begin{table*}[!ht]
\centering
\resizebox{0.99\textwidth}{!}{%
\begin{tabular}{lcccccccc}
\toprule
\multirow{3}{*}{Agent} & \multirow{3}{*}{Base Model} & \multirow{3}{*}{\begin{tabular}[c]{@{}c@{}}SR$\uparrow$\end{tabular}} & \multicolumn{3}{c}{Efficiency} & \multicolumn{2}{c}{Cost} & \multirow{3}{*}{\begin{tabular}[c]{@{}c@{}}S2GR$\uparrow$\end{tabular}} \\
\cmidrule(lr){4-6} \cmidrule(lr){7-8}
    &  &  & \begin{tabular}[c]{@{}c@{}}MS$\downarrow$\end{tabular} & \begin{tabular}[c]{@{}c@{}}MSRS$\downarrow$\end{tabular} & \begin{tabular}[c]{@{}c@{}}MET$\downarrow$\end{tabular} & \begin{tabular}[c]{@{}c@{}}MToC$\downarrow$\end{tabular} & \begin{tabular}[c]{@{}c@{}}MSC$\uparrow$\end{tabular} &  \\
\midrule
\rowcolor[HTML]{EFEFEF}
\multicolumn{9}{c}{\textit{Single-app Tasks (92 Tasks)}} \\
\midrule
\textit{Human} & - & - & \textit{9.272} & \textit{1.000} & - & - & - & - \\
\midrule
MobileAgentV2 & Gemini-2.0-Flash & 0.250 & 12.881 & 1.280 & \underline{472.426} & \underline{40.406} & 0 & 0 \\
MAS-MobileAgent & Gemini-2.0-Flash & 0.402 & \underline{9.087}{\small\textcolor[RGB]{0,128,0}{$_{\uparrow29\%}$}} & \underline{0.719}{\small\textcolor[RGB]{0,128,0}{$_{\uparrow44\%}$}} & \textbf{335.181} & \textbf{39.597} & \textbf{3.565} & \textbf{0.505} \\
\cmidrule(lr){1-9}
MobileAgentV2 & Gemini-2.5-Pro & \underline{0.446} & 12.424 & 1.058 & 1013.386 & 120.212 & 0 & 0 \\
MAS-MobileAgent & Gemini-2.5-Pro & \textbf{0.641} & \textbf{9.109}{\small\textcolor[RGB]{0,128,0}{$_{\uparrow27\%}$}} & \textbf{0.613}{\small\textcolor[RGB]{0,128,0}{$_{\uparrow42\%}$}} & 682.547 & 99.780 & 1.348 & 0.345 \\
\midrule
\rowcolor[HTML]{EFEFEF}
\multicolumn{9}{c}{\textit{Cross-app Tasks (47 Tasks)}} \\
\midrule
\textit{Human} & - & - & \textit{17.660} & \textit{1.000} & - & - & - & - \\
\midrule
MobileAgentV2 & Gemini-2.0-Flash & 0 & 31.553 & - & \underline{952.660}  & \underline{106.501} & 0 & 0 \\
MAS-MobileAgent & Gemini-2.0-Flash & \underline{0.234} & \underline{24.132}{\small\textcolor[RGB]{0,128,0}{$_{\uparrow24\%}$}} & \underline{0.938} & \textbf{403.798} & \textbf{71.350} & \textbf{5.362} & \textbf{0.709} \\
\cmidrule(lr){1-9}
MobileAgentV2 & Gemini-2.5-Pro & 0.170 & 32.617 & 1.247 & 2053.133 & 227.128 & 0 & 0 \\
MAS-MobileAgent & Gemini-2.5-Pro & \textbf{0.617} & \textbf{20.404}{\small\textcolor[RGB]{0,128,0}{$_{\uparrow37\%}$}} & \textbf{0.829}{\small\textcolor[RGB]{0,128,0}{$_{\uparrow34\%}$}} & 1441.586 & 189.836 & 3.128 & 0.320 \\
\bottomrule
\end{tabular}
}
\caption{Evaluation results of different base models. MS: Mean Steps; MSRS: Mean Step Ratio on Successful tasks; MET: Mean Execution Time (seconds); MToC: Mean Token Cost (in thousands); MSC: Mean Shortcut Call count; S2GR: Shortcut-to-GUI action Ratio.}
\label{table:flash_diff_pro}
\end{table*}

\subsection{Details of Tasks in MAS-Bench}
\label{app:task_design}

We have developed a specialized benchmark suite to validate the feasibility of GUI-shortcut hybrid actions for mobile agents and establish a framework for the comprehensive evaluation of an agent's ability to utilize and generate shortcuts.

This suite comprises 139 tasks strategically designed across 11 popular real-world applications. The tasks are categorized into 92 single-app tasks and 47 cross-app tasks, with their difficulty distribution illustrated in Table~\ref{table:difficulty_level}. We have also curated a corresponding knowledge base of predefined shortcuts to support these tasks. Table~\ref{table:app-shortcut} presents a breakdown of the selected applications and the number of single-app tasks and shortcuts associated with each.

Table~\ref{tab:task_category_examples} summarizes the task coverage and representative examples. Single-app tasks cover shopping, local services, information browsing, media, productivity, health, and navigation workflows. Approximately 37\% of single-app tasks are designed as incremental variations, where later tasks extend earlier ones with additional constraints or sub-goals. Cross-app tasks emphasize inter-app coordination and are mostly constructed by composing foundational single-app operations into information transfer, sharing, planning, and multi-app workflows.

\begin{table*}[t]
\centering
\resizebox{0.98\textwidth}{!}{%
\begin{tabular}{llclp{6.6cm}}
\toprule
Task Type & Category & \# Tasks & Representative Apps & Example Task Pattern \\
\midrule
Single-app & Shopping and local services & 28 & Amazon, Booking.com, Yelp & Search, filter, and verify products, hotels, attractions, or restaurants within one app. \\
Single-app & Information and media browsing & 26 & BBC News, Chrome, YouTube & Navigate news sections, search the web or videos, and save or play selected content. \\
Single-app & Productivity, health, and navigation & 38 & Contacts, Fitbit, Gmail, Calendar, Maps & Create contacts, draft emails, add calendar events, log health records, or obtain map directions. \\
\midrule
Cross-app & Information management & 16 & YouTube, Calendar, Booking.com, Contacts, BBC News & Retrieve information in one app and record, schedule, or reuse it in another app. \\
Cross-app & Web shopping and travel planning & 11 & Amazon, Booking.com, Gmail, Calendar, Chrome & Search products, hotels, flights, or rentals and transfer the result to email, calendar, or another app. \\
Cross-app & Multi-app coordination & 11 & Maps, Yelp, Gmail, Calendar, BBC News & Coordinate several apps to complete chained goals such as finding places and sharing details. \\
Cross-app & Social sharing & 5 & Maps, BBC News, Gmail & Find content such as locations, photos, or articles and share the result through email. \\
Cross-app & Media and entertainment & 4 & YouTube, Chrome, Fitbit, Calendar & Search or consume media content and connect it with logging or scheduling actions. \\
\bottomrule
\end{tabular}
}
\caption{\textbf{Task categories and representative examples in MAS-Bench.} MAS-Bench contains 92 single-app tasks and 47 cross-app tasks across 11 real-world Android applications.}
\label{tab:task_category_examples}
\end{table*}

\paragraph{Human Baseline Collection Methodology.}
To establish optimal step counts for efficiency metrics (MSR and MSRS), we conducted a systematic human expert annotation process. Three experienced mobile device users were recruited to independently complete each of the 139 tasks in MAS-Bench. Each expert was instructed to complete tasks as efficiently as possible using \textbf{only pure GUI operations}, specifically standard interactions such as taps, swipes, and text input, without access to any programmatic shortcuts. For each task, we recorded the number of steps taken by each expert and selected the \textit{minimum} step count across the three demonstrations as the human baseline. This methodology ensures that our baseline represents the shortest achievable pure-GUI path under realistic conditions, providing a fair and reproducible reference for agent efficiency evaluation. The resulting human baselines average 9.27 steps for single-app tasks and 17.66 steps for cross-app tasks, reflecting the inherent complexity difference between these task categories.

\begin{table}[h]
\centering
\resizebox{0.99\linewidth}{!}{%
\begin{tabular}{lcc}
\toprule
\textbf{Difficulty Level} & \textbf{Proportion (\%)} & \textbf{Human Steps} \\
\midrule
Level 1 & 27.3 & 6.2\\
Level 2 & 47.5 & 12.6\\
Level 3 & 25.2 & 17.6\\
\bottomrule
\end{tabular}
}
\caption{Distribution of task difficulty levels in MAS-Bench. The table shows the proportion of tasks at each level and the average number of steps required for a human to complete them.}
\label{table:difficulty_level}
\end{table}

\subsection{Evaluation Metrics}
\label{app:metrics_summary}

To comprehensively evaluate agent performance on MAS-Bench, we employ a multi-dimensional evaluation framework that assesses success, efficiency, and cost. Table~\ref{table:metrics_summary} provides a complete summary of all metrics. Our evaluation framework is designed to encourage agents to efficiently complete tasks by intelligently utilizing existing and self-generated shortcuts, thereby achieving lower costs and greater efficiency.

\paragraph{Success Rate (SR).} We use SR to measure whether an agent completes a task. Success is typically defined as the agent 
reaching the final goal state of the task while satisfying all of its requirements.

\paragraph{Efficiency.} We use the following metrics to evaluate the impact of shortcut actions on agents' efficiency, measured 
by the time and operations required to complete a task:

\begin{itemize}
    \item \textbf{Mean Steps (MS)}: The average number of steps an agent takes to complete a task. 
    \item \textbf{Mean Step Ratio (MSR)}: The ratio of agent steps to human steps (pure GUI operations). Lower values indicate better efficiency, with values below 1.0 representing fewer steps than the human.
    \item \textbf{Mean Step Ratio on Successful tasks (MSRS)}: The same ratio as MSR, computed only over successfully completed tasks.
    \item \textbf{Mean Execution Time (MET)}: The average time an agent takes to complete a task.
\end{itemize}

\paragraph{Cost and Resource Utilization.} 
Cost metrics evaluate the computational resources and overhead an agent consumes to complete a task. A high-performing agent 
should minimize the overall cost by intelligently utilizing or autonomously generating shortcuts. We use the following metrics to 
evaluate the cost of an agent:

\begin{itemize}
    \item \textbf{Mean Token Cost (MToC)}: The average number of tokens (in thousands) consumed by the LLM during task execution. 
    \item \textbf{Mean Shortcut Call Count (MSC)}: The average number of shortcuts the agent calls during the task execution.
    \item \textbf{Shortcut Success Rate (SSR)}: The ratio of successfully executed shortcut calls to the total number of shortcut 
    calls made by the agent.
    \item \textbf{Shortcut to GUI Action Ratio (S2GR)}: This metric reflects an agent's operational strategy preference by 
    calculating the ratio of shortcut operations to GUI operations. A higher ratio indicates that the agent relies more heavily 
    on efficient shortcuts rather than manual GUI actions.
\end{itemize}

\begin{table*}[t]
\centering
\small
\begin{tabular}{llp{7cm}cc}
\toprule
\textbf{Metric} & \textbf{Symbol} & \textbf{Definition} & \textbf{Unit} & \textbf{Dir.} \\
\midrule
\multicolumn{5}{c}{\textit{Success}} \\
\midrule
Success Rate & SR & Percentage of successfully completed tasks & \% & $\uparrow$ \\
\midrule
\multicolumn{5}{c}{\textit{Efficiency}} \\
\midrule
Mean Steps & MS & Average number of steps (GUI actions + shortcuts) per task & steps & $\downarrow$ \\
Mean Step Ratio & MSR & Ratio of agent steps to human baseline steps (pure GUI) & ratio & $\downarrow$ \\
Mean Step Ratio on Successful & MSRS & MSR calculated only on successfully completed tasks & ratio & $\downarrow$ \\
Mean Execution Time & MET & Average wall-clock time to complete a task & seconds & $\downarrow$ \\
\midrule
\multicolumn{5}{c}{\textit{Cost and Resource Utilization}} \\
\midrule
Mean Token Cost & MToC & Average LLM tokens consumed per task & thousands & $\downarrow$ \\
Mean Shortcut Call Count & MSC & Average number of shortcut invocations per task & count & $\uparrow$ \\
Shortcut Success Rate & SSR & Ratio of successful shortcut calls to total calls & \% & $\uparrow$ \\
Shortcut-to-GUI Ratio & S2GR & Ratio of shortcut actions to GUI actions & ratio & $\uparrow$ \\
\bottomrule
\end{tabular}
\caption{\textbf{Summary of evaluation metrics in MAS-Bench.} Dir.: Direction of improvement ($\uparrow$ higher is better, $\downarrow$ lower is better). MSR and MSRS use human expert demonstrations (pure GUI operations) as the baseline. Values below 1.0 indicate agent performance exceeds human efficiency through shortcut utilization.}
\label{table:metrics_summary}
\end{table*}

\subsection{Task Difficulty Classification}
\label{app:difficulty_classification}

MAS-Bench tasks are categorized into three difficulty levels to enable systematic evaluation across varying complexity. The classification criteria differ between single-app and cross-app tasks, reflecting their distinct operational characteristics.

\paragraph{Single-App Tasks.} For the 92 single-app tasks, difficulty is primarily determined by \textbf{operational complexity} measured through the number of required steps and the sophistication of sub-goals:

\begin{itemize}
    \item \textbf{Level 1 (Easy)}: Tasks requiring $\leq 7$ steps. These involve basic, direct operations with a single primary goal, such as simple searches or navigation (e.g., searching for a product, opening a specific page).
    
    \item \textbf{Level 2 (Medium)}: Tasks requiring 8-15 steps. These involve multi-step workflows with intermediate complexity, including filtering operations, state verification, or coordination of multiple sub-goals (e.g., searching with filters and adding items to cart).
    
    \item \textbf{Level 3 (Hard)}: Tasks requiring $\geq 15$ steps. These involve complex workflows with multiple interdependent sub-goals, advanced filtering, cross-page navigation, and explicit state verification (e.g., multi-criteria product comparison with cart management and confirmation).
\end{itemize}

\paragraph{Cross-App Tasks.} For the 47 cross-app tasks, difficulty is determined by both the \textbf{number of applications involved} and the \textbf{complexity of inter-app coordination}:

\begin{itemize}
    \item \textbf{Level 1 (Easy)}: Tasks involving 2 applications with straightforward information transfer, typically requiring $\leq 12$ steps. These tasks involve basic data lookup in one app and simple recording or sharing in another.
    
    \item \textbf{Level 2 (Medium)}: Tasks involving 2-3 applications with moderate coordination complexity, typically requiring 13-20 steps. These tasks involve filtering, processing, or transforming information across applications.
    
    \item \textbf{Level 3 (Hard)}: Tasks involving 3-4 applications with complex multi-app workflows, typically requiring $\geq 20$ steps. These tasks demand sophisticated inter-app dependencies, multi-stage data processing, and coordination across diverse application domains.
\end{itemize}

\section{Details of Predefined Shortcut Evaluation}

\subsection{Detailed Results on Single-app and Cross-app Tasks}
\label{app:detailed_results}

This section provides detailed performance for representative agents on single-app and cross-app tasks in MAS-Bench. Table~\ref{tab:results-masbench-single} shows the results for 92 single-app tasks, while Table~\ref{tab:results-masbench-cross} presents the results for 47 cross-app tasks. These detailed results complement the overall weighted average results shown in Table~\ref{tab:results-masbench} in the main text.

\begin{table*}[!ht]
\centering
\resizebox{0.99\textwidth}{!}{%
\begin{tabular}{lccccccccc}
\toprule
\multirow{3}{*}{Agent} & \multicolumn{2}{c}{Input} & \multirow{3}{*}{\begin{tabular}[c]{@{}c@{}}SR$\uparrow$\end{tabular}} & \multicolumn{3}{c}{Efficiency} & \multicolumn{2}{c}{Cost} & \multirow{3}{*}{\begin{tabular}[c]{@{}c@{}}S2GR$\uparrow$\end{tabular}} \\
\cmidrule(lr){2-3} \cmidrule(lr){5-7} \cmidrule(lr){8-9}
    & \begin{tabular}[c]{@{}c@{}}SS\end{tabular} & \begin{tabular}[c]{@{}c@{}}VH\end{tabular} & & \begin{tabular}[c]{@{}c@{}}MS$\downarrow$\end{tabular} & \begin{tabular}[c]{@{}c@{}}MSRS$\downarrow$\end{tabular} & \begin{tabular}[c]{@{}c@{}}MET$\downarrow$\end{tabular} & \begin{tabular}[c]{@{}c@{}}MToC$\downarrow$\end{tabular} & \begin{tabular}[c]{@{}c@{}}MSC$\uparrow$\end{tabular} &  \\
\midrule
\textit{Human}  & \checkmarkmy &  & - & \textit{9.272} & \textit{1.000} & - & - & - & - \\
\midrule
\rowcolor[HTML]{E8E8E8}
\multicolumn{10}{c}{\textit{Agentic Workflow (Gemini-2.5-Pro)}} \\
\midrule
M3A~\cite{rawles2024androidworld} & \checkmarkmy & \checkmarkmy & 0.565 & 11.772 & 1.064 & 192.775 & 155.281 & 0 & 0 \\
MobileAgent-E~\cite{wang2025mobile} & \checkmarkmy &  & 0.359 & \textbf{4.080} & \underline{0.818} & 459.574 & \textbf{88.772} & 0.378 & 0.081 \\
\cmidrule(lr){1-10}
T3A~\cite{rawles2024androidworld} & & \checkmarkmy  & 0.511 & 11.750 & 1.056 & \underline{137.641} & 346.382 & 0 & 0 \\
\rowcolor[HTML]{E6F2FF}
\quad\textbf{+ MAS-T3A (Ours)} & & \checkmarkmy & \underline{0.576}{\small\textcolor{blue}{$_{\uparrow13\%}$}} & 10.595{\small\textcolor{blue}{$_{\uparrow10\%}$}} & 0.915{\small\textcolor{blue}{$_{\uparrow13\%}$}} & \textbf{129.279}{\small\textcolor{blue}{$_{\uparrow6\%}$}} & 291.391{\small\textcolor{blue}{$_{\uparrow16\%}$}} & \underline{1.043} & \underline{0.117} \\
\cmidrule(lr){1-10}
MobileAgentV2~\cite{wang2024mobileagentv2} & \checkmarkmy &  & 0.446 & 12.424 & 1.058 & 1013.386 & 120.212 & 0 & 0 \\
\rowcolor[HTML]{E6F2FF}
\quad\textbf{+ MAS-MobileAgent (Ours)} & \checkmarkmy &  & \textbf{0.641}{\small\textcolor{blue}{$_{\uparrow44\%}$}} & \underline{9.109}{\small\textcolor{blue}{$_{\uparrow27\%}$}} & \textbf{0.613}{\small\textcolor{blue}{$_{\uparrow42\%}$}} & 682.547{\small\textcolor{blue}{$_{\uparrow33\%}$}} & \underline{99.780}{\small\textcolor{blue}{$_{\uparrow17\%}$}} & \textbf{1.348} & \textbf{0.345} \\
\midrule
\rowcolor[HTML]{E8E8E8}
\multicolumn{10}{c}{\textit{General-Purpose Models}} \\
\midrule
Qwen2.5-VL-3B~\cite{bai2025qwen2} & \checkmarkmy &  & 0.054 & 17.380 & 1.341 & \textbf{99.564} & - & 0 & 0 \\
Qwen2.5-VL-7B~\cite{bai2025qwen2} & \checkmarkmy &  & 0.022 & 17.315 & 1.625 & 167.892 & - & 0 & 0 \\
\cmidrule(lr){1-10}
Qwen3-VL-4B~\cite{bai2025qwen3} & \checkmarkmy &  & 0.290 & 13.739 & 1.274 & 143.098 & - & 0 & 0 \\
\rowcolor[HTML]{E6F2FF}
\quad\textbf{+ MAS-Qwen3-VL-4B (Ours)} & \checkmarkmy &  & 0.315{\small\textcolor{blue}{$_{\uparrow8.6\%}$}} & 15.011{\small\textcolor{red}{$_{-9.3\%}$}} & 0.888{\small\textcolor{blue}{$_{\uparrow30.3\%}$}} & 147.230{\small\textcolor{red}{$_{-2.9\%}$}} & - & \textbf{2.620} & \underline{0.219} \\
\cmidrule(lr){1-10}
Qwen3-VL-8B~\cite{bai2025qwen3} & \checkmarkmy &  & 0.326 & 12.761 & 1.179 & 148.021 & - & 0 & 0 \\
\rowcolor[HTML]{E6F2FF}
\quad\textbf{+ MAS-Qwen3-VL-8B (Ours)} & \checkmarkmy &  & 0.522{\small\textcolor{blue}{$_{\uparrow60\%}$}} & 11.065{\small\textcolor{blue}{$_{\uparrow13\%}$}} & \textbf{0.792}{\small\textcolor{blue}{$_{\uparrow33\%}$}} & \underline{124.370}{\small\textcolor{blue}{$_{\uparrow16\%}$}} & - & 1.239 & 0.129 \\
\cmidrule(lr){1-10}
Qwen3-VL-32B~\cite{bai2025qwen3} & \checkmarkmy &  & 0.402 & 12.793 & 1.150 & 293.941 & - & 0 & 0 \\
\rowcolor[HTML]{E6F2FF}
\quad\textbf{+ MAS-Qwen3-VL-32B (Ours)} & \checkmarkmy &  & 0.543{\small\textcolor{blue}{$_{\uparrow35.1\%}$}} & 11.859{\small\textcolor{blue}{$_{\uparrow7.3\%}$}} & \underline{0.810}{\small\textcolor{blue}{$_{\uparrow29.6\%}$}} & 269.291{\small\textcolor{blue}{$_{\uparrow8.4\%}$}} & - & \underline{2.272} & \textbf{0.271} \\
\cmidrule(lr){1-10}
Qwen3-VL-235B~\cite{bai2025qwen3} & \checkmarkmy &  & 0.478 & 11.989 & 1.113 & 166.083 & - & 0 & 0 \\
\rowcolor[HTML]{E6F2FF}
\quad\textbf{+ MAS-Qwen3-VL-235B (Ours)} & \checkmarkmy &  & \underline{0.598}{\small\textcolor{blue}{$_{\uparrow25.1\%}$}} & \underline{10.924}{\small\textcolor{blue}{$_{\uparrow8.9\%}$}} & 0.849{\small\textcolor{blue}{$_{\uparrow23.7\%}$}} & 134.274{\small\textcolor{blue}{$_{\uparrow19.2\%}$}} & - & 1.989 & \underline{0.219} \\
\cmidrule(lr){1-10}
GLM-4.5V~\cite{hong2025glm} & \checkmarkmy &  & 0.533 & 13.739 & 1.209 & 214.544 & - & 0 & 0 \\
\rowcolor[HTML]{E6F2FF}
\quad\textbf{+ MAS-GLM-4.5V (Ours)} & \checkmarkmy &  & \textbf{0.739}{\small\textcolor{blue}{$_{\uparrow38.6\%}$}} & \textbf{10.337}{\small\textcolor{blue}{$_{\uparrow24.8\%}$}} & 0.823{\small\textcolor{blue}{$_{\uparrow31.9\%}$}} & 192.489{\small\textcolor{blue}{$_{\uparrow10.3\%}$}} & - & 0.957 & 0.110 \\
\midrule
\rowcolor[HTML]{E8E8E8}
\multicolumn{10}{c}{\textit{Specialized GUI Models}} \\
\midrule
UI-TARS-1.5-7B~\cite{ui-tars-15} & \checkmarkmy &  & 0.380 & 14.543 & 1.203 & 147.966 & - & 0 & 0 \\
GUI-Owl-7B~\cite{mobileagentv3} & \checkmarkmy &  & 0.402 & \textbf{11.554} & 1.279 & 132.698 & - & 0 & 0 \\
\cmidrule(lr){1-10}
ScaleCUA-7B~\cite{liu2025scalecua} & \checkmarkmy &  & 0.163 & 13.554 & \underline{1.202} & \textbf{98.037} & - & 0 & 0 \\
\rowcolor[HTML]{E6F2FF}
\quad\textbf{+ MAS-ScaleCUA-7B (Ours)} & \checkmarkmy &  & 0.174{\small\textcolor{blue}{$_{\uparrow6.7\%}$}} & 14.685{\small\textcolor{red}{$_{-8.3\%}$}} & 1.358{\small\textcolor{red}{$_{-13.0\%}$}} & \underline{111.684}{\small\textcolor{red}{$_{-13.9\%}$}} & - & \underline{0.011} & \underline{0.001} \\
\cmidrule(lr){1-10}
ScaleCUA-32B~\cite{liu2025scalecua} & \checkmarkmy &  & 0.283 & 14.065 & 1.273 & 113.134 & - & 0 & 0 \\
\rowcolor[HTML]{E6F2FF}
\quad\textbf{+ MAS-ScaleCUA-32B (Ours)} & \checkmarkmy &  & 0.272{\small\textcolor{red}{$_{-3.9\%}$}} & 14.230{\small\textcolor{red}{$_{-1.2\%}$}} & 1.245{\small\textcolor{blue}{$_{\uparrow2.2\%}$}} & 115.300{\small\textcolor{red}{$_{-1.9\%}$}} & - & 0 & 0 \\
\cmidrule(lr){1-10}
MAI-UI-8B~\cite{zhou2025mai} & \checkmarkmy &  & \underline{0.565} & 13.793 & 1.213 & 137.677 & - & 0 & 0 \\
\rowcolor[HTML]{E6F2FF}
\quad\textbf{+ MAS-MAI-UI-8B (Ours)} & \checkmarkmy &  & \textbf{0.652}{\small\textcolor{blue}{$_{\uparrow15.4\%}$}} & \underline{13.065}{\small\textcolor{blue}{$_{\uparrow5.3\%}$}} & \textbf{1.130}{\small\textcolor{blue}{$_{\uparrow6.8\%}$}} & 131.573{\small\textcolor{blue}{$_{\uparrow4.4\%}$}} & - & \textbf{0.315} & \textbf{0.024} \\
\bottomrule
\end{tabular}
}
\caption{\textbf{Detailed performance on single-app tasks (92 tasks).} Bold and underlined values denote the best and second-best results within each block, respectively. Methods marked with ``\textbf{+}'' are shortcut-augmented variants of their corresponding baselines. SS: Screenshot; VH: View Hierarchy; MS: Mean Steps; MSRS: Mean Step Ratio on Successful tasks; MET: Mean Execution Time (seconds); MToC: Mean Token Cost (in thousands); MSC: Mean Shortcut Call count; S2GR: Shortcut-to-GUI action Ratio.}
\label{tab:results-masbench-single}
\end{table*}

\begin{table*}[!ht]
\centering
\resizebox{0.99\textwidth}{!}{%
\begin{tabular}{lccccccccc}
\toprule
\multirow{3}{*}{Agent} & \multicolumn{2}{c}{Input} & \multirow{3}{*}{\begin{tabular}[c]{@{}c@{}}SR$\uparrow$\end{tabular}} & \multicolumn{3}{c}{Efficiency} & \multicolumn{2}{c}{Cost} & \multirow{3}{*}{\begin{tabular}[c]{@{}c@{}}S2GR$\uparrow$\end{tabular}} \\
\cmidrule(lr){2-3} \cmidrule(lr){5-7} \cmidrule(lr){8-9}
    & \begin{tabular}[c]{@{}c@{}}SS\end{tabular} & \begin{tabular}[c]{@{}c@{}}VH\end{tabular} & & \begin{tabular}[c]{@{}c@{}}MS$\downarrow$\end{tabular} & \begin{tabular}[c]{@{}c@{}}MSRS$\downarrow$\end{tabular} & \begin{tabular}[c]{@{}c@{}}MET$\downarrow$\end{tabular} & \begin{tabular}[c]{@{}c@{}}MToC$\downarrow$\end{tabular} & \begin{tabular}[c]{@{}c@{}}MSC$\uparrow$\end{tabular} &  \\
\midrule
\textit{Human} & \checkmarkmy &  & - & \textit{17.660} & \textit{1.000} & - & - & - & - \\
\midrule
\rowcolor[HTML]{E8E8E8}
\multicolumn{10}{c}{\textit{Agentic Workflow (Gemini-2.5-Pro)}} \\
\midrule
M3A~\cite{rawles2024androidworld} & \checkmarkmy & \checkmarkmy & 0.383 & 26.213 & 1.262 & 410.828 & 289.040 & 0 & 0 \\
MobileAgent-E~\cite{wang2025mobile} & \checkmarkmy &  & 0.064 & \textbf{6.234} & 0.934 & 468.630 & \textbf{85.954} & \underline{2.250} & 0.177 \\
\cmidrule(lr){1-10}
T3A~\cite{rawles2024androidworld} & & \checkmarkmy & 0.340 & 23.596 & 1.087 & \underline{257.015} & 625.970 & 0 & 0 \\
\rowcolor[HTML]{E6F2FF}
\quad\textbf{+ MAS-T3A (Ours)} & &\checkmarkmy & \underline{0.511}{\small\textcolor{blue}{$_{\uparrow50\%}$}} & \underline{17.021}{\small\textcolor{blue}{$_{\uparrow28\%}$}} & \textbf{0.643}{\small\textcolor{blue}{$_{\uparrow41\%}$}} & \textbf{186.139}{\small\textcolor{blue}{$_{\uparrow28\%}$}} & 439.296{\small\textcolor{blue}{$_{\uparrow30\%}$}} & 2.213 & \underline{0.185} \\
\cmidrule(lr){1-10}
MobileAgentV2~\cite{wang2024mobileagentv2} & \checkmarkmy &  & 0.170 & 32.617 & 1.247 & 2053.204 & 227.357 & 0 & 0 \\
\rowcolor[HTML]{E6F2FF}
\quad\textbf{+ MAS-MobileAgent (Ours)} & \checkmarkmy &  & \textbf{0.617}{\small\textcolor{blue}{$_{\uparrow263\%}$}} & 20.404{\small\textcolor{blue}{$_{\uparrow37\%}$}} & \underline{0.829}{\small\textcolor{blue}{$_{\uparrow34\%}$}} & 1479.197{\small\textcolor{blue}{$_{\uparrow28\%}$}} & \underline{192.052}{\small\textcolor{blue}{$_{\uparrow16\%}$}} & \textbf{3.122} & \textbf{0.333} \\
\midrule
\rowcolor[HTML]{E8E8E8}
\multicolumn{10}{c}{\textit{General-Purpose Models}} \\
\midrule
Qwen2.5-VL-3B~\cite{bai2025qwen2} & \checkmarkmy &  & 0 & 29.936 & - & \textbf{162.886} & - & 0 & 0 \\
Qwen2.5-VL-7B~\cite{bai2025qwen2} & \checkmarkmy &  & 0.021 & 30.681 & 1.250 & \underline{170.672} & - & 0 & 0 \\
\cmidrule(lr){1-10}
Qwen3-VL-4B~\cite{bai2025qwen3} & \checkmarkmy &  & 0.106 & 25.000 & 1.272 & 249.891 & - & 0 & 0 \\
\rowcolor[HTML]{E6F2FF}
\quad\textbf{+ MAS-Qwen3-VL-4B (Ours)} & \checkmarkmy &  & 0.085{\small\textcolor{red}{$_{-19.7\%}$}} & 30.957{\small\textcolor{red}{$_{-23.8\%}$}} & 1.133{\small\textcolor{blue}{$_{\uparrow10.9\%}$}} & 268.035{\small\textcolor{red}{$_{-7.3\%}$}} & - & 2.149 & 0.075 \\
\cmidrule(lr){1-10}
Qwen3-VL-8B~\cite{bai2025qwen3} & \checkmarkmy &  & 0.128 & 24.043 & 1.192 & 240.755 & - & 0 & 0 \\
\rowcolor[HTML]{E6F2FF}
\quad\textbf{+ MAS-Qwen3-VL-8B (Ours)} & \checkmarkmy &  & 0.234{\small\textcolor{blue}{$_{\uparrow83\%}$}} & 22.468{\small\textcolor{blue}{$_{\uparrow7\%}$}} & 1.016{\small\textcolor{blue}{$_{\uparrow15\%}$}} & 217.846{\small\textcolor{blue}{$_{\uparrow9\%}$}} & - & 1.447 & 0.073 \\
\cmidrule(lr){1-10}
Qwen3-VL-32B~\cite{bai2025qwen3} & \checkmarkmy &  & 0.213 & 27.702 & 1.318 & 590.161 & - & 0 & 0 \\
\rowcolor[HTML]{E6F2FF}
\quad\textbf{+ MAS-Qwen3-VL-32B (Ours)} & \checkmarkmy &  & 0.255{\small\textcolor{blue}{$_{\uparrow19.7\%}$}} & 23.849{\small\textcolor{blue}{$_{\uparrow13.9\%}$}} & \underline{0.923}{\small\textcolor{blue}{$_{\uparrow30.0\%}$}} & 532.476{\small\textcolor{blue}{$_{\uparrow9.8\%}$}} & - & \textbf{5.064} & \textbf{0.287} \\
\cmidrule(lr){1-10}
Qwen3-VL-235B~\cite{bai2025qwen3} & \checkmarkmy &  & 0.298 & 25.638 & 1.106 & 222.624 & - & 0 & 0 \\
\rowcolor[HTML]{E6F2FF}
\quad\textbf{+ MAS-Qwen3-VL-235B (Ours)} & \checkmarkmy &  & 0.383{\small\textcolor{blue}{$_{\uparrow28.5\%}$}} & \textbf{21.276}{\small\textcolor{blue}{$_{\uparrow17.0\%}$}} & \textbf{0.659}{\small\textcolor{blue}{$_{\uparrow40.4\%}$}} & 188.626{\small\textcolor{blue}{$_{\uparrow15.3\%}$}} & - & \underline{4.340} & \underline{0.261} \\
\cmidrule(lr){1-10}
GLM-4.5V~\cite{hong2025glm} & \checkmarkmy &  & \underline{0.511} & 24.085 & 1.165 & 413.829 & - & 0 & 0 \\
\rowcolor[HTML]{E6F2FF}
\quad\textbf{+ MAS-GLM-4.5V (Ours)} & \checkmarkmy &  & \textbf{0.574}{\small\textcolor{blue}{$_{\uparrow12.3\%}$}} & \underline{21.319}{\small\textcolor{blue}{$_{\uparrow11.5\%}$}} & 1.051{\small\textcolor{blue}{$_{\uparrow9.8\%}$}} & 328.798{\small\textcolor{blue}{$_{\uparrow20.5\%}$}} & - & 1.234 & 0.071 \\
\midrule
\rowcolor[HTML]{E8E8E8}
\multicolumn{10}{c}{\textit{Specialized GUI Models}} \\
\midrule
UI-TARS-1.5-7B~\cite{ui-tars-15} & \checkmarkmy &  & 0.106 & 28.340 & \underline{1.169} & 266.787 & - & 0 & 0 \\
GUI-Owl-7B~\cite{mobileagentv3} & \checkmarkmy &  & 0.085 & \textbf{23.426} & \textbf{1.161} & 237.539 & - & 0 & 0 \\
\cmidrule(lr){1-10}
ScaleCUA-7B~\cite{liu2025scalecua} & \checkmarkmy &  & 0 & \underline{27.277} & - & \textbf{171.917} & - & 0 & 0 \\
\rowcolor[HTML]{E6F2FF}
\quad\textbf{+ MAS-ScaleCUA-7B (Ours)} & \checkmarkmy &  & 0 & 28.766{\small\textcolor{red}{$_{-5.5\%}$}} & - & 199.968{\small\textcolor{red}{$_{-16.3\%}$}} & - & 0 & 0 \\
\cmidrule(lr){1-10}
ScaleCUA-32B~\cite{liu2025scalecua} & \checkmarkmy &  & 0.128 & 29.277 & 1.310 & 197.481 & - & 0 & 0 \\
\rowcolor[HTML]{E6F2FF}
\quad\textbf{+ MAS-ScaleCUA-32B (Ours)} & \checkmarkmy &  & 0.106{\small\textcolor{red}{$_{-17.2\%}$}} & 28.145{\small\textcolor{blue}{$_{\uparrow3.9\%}$}} & 1.290{\small\textcolor{blue}{$_{\uparrow1.5\%}$}} & \underline{188.723}{\small\textcolor{blue}{$_{\uparrow4.4\%}$}} & - & 0 & 0 \\
\cmidrule(lr){1-10}
MAI-UI-8B~\cite{zhou2025mai} & \checkmarkmy &  & \underline{0.340} & 29.894 & 1.313 & 264.849 & - & 0 & 0 \\
\rowcolor[HTML]{E6F2FF}
\quad\textbf{+ MAS-MAI-UI-8B (Ours)} & \checkmarkmy &  & \textbf{0.447}{\small\textcolor{blue}{$_{\uparrow31.5\%}$}} & 27.851{\small\textcolor{blue}{$_{\uparrow6.8\%}$}} & 1.274{\small\textcolor{blue}{$_{\uparrow3.0\%}$}} & 243.632{\small\textcolor{blue}{$_{\uparrow8.0\%}$}} & - & \textbf{0.191} & \textbf{0.007} \\
\bottomrule
\end{tabular}
}
\caption{\textbf{Detailed performance on cross-app tasks (47 tasks).} Bold and underlined values denote the best and second-best results within each block, respectively. Methods marked with ``\textbf{+}'' are shortcut-augmented variants of their corresponding baselines. SS: Screenshot; VH: View Hierarchy; MS: Mean Steps; MSRS: Mean Step Ratio on Successful tasks; MET: Mean Execution Time (seconds); MToC: Mean Token Cost (in thousands); MSC: Mean Shortcut Call count; S2GR: Shortcut-to-GUI action Ratio.}
\label{tab:results-masbench-cross}
\end{table*}

\subsubsection{Detailed Results of Single-app and Cross-app Tasks}

The performance comparison between single-app and cross-app tasks reveals several critical insights into agent capabilities and the effectiveness of shortcut augmentation. Table~\ref{tab:results-masbench-single} and Table~\ref{tab:results-masbench-cross} show the detailed results for representative agents.

\paragraph{Task Complexity and Performance Degradation.}
Cross-app tasks demonstrate significantly higher complexity than single-app tasks, as evidenced by the substantial performance degradation across many agents. For instance, the baseline T3A achieves an SR of 51.1\% on single-app tasks but drops to 34.0\% on cross-app tasks, representing a 33\% relative decrease. Similarly, MobileAgentV2 shows an even more pronounced decline from 44.6\% to 17.0\% (62\% relative decrease). This trend is also observed in several Agent-as-a-Model baselines, with GUI-Owl-7B declining from 40.2\% to 8.5\%. The human baseline steps also increase from 9.272 to 17.660, confirming the inherent complexity difference between task types.

\paragraph{Amplified Benefits of Shortcuts on Cross-app Tasks.}
Predefined shortcuts demonstrate disproportionately larger benefits on cross-app tasks compared to single-app tasks. MAS-MobileAgent improves MobileAgentV2's SR by 44\% on single-app tasks but achieves a remarkable 263\% improvement on cross-app tasks (17.0\% → 61.7\%). Similarly, MAS-T3A shows a 13\% improvement on single-app tasks versus 50\% on cross-app tasks. This amplified effect stems from shortcuts' ability to bridge cross-application boundaries. Complex inter-app transitions that are particularly challenging for GUI-only agents become single atomic actions with appropriate shortcuts.

\paragraph{Efficiency Gains Across Task Types.}
The Mean Step Ratio on Successful tasks (MSRS) metric reveals consistent efficiency improvements from shortcut augmentation across both task types. For agentic workflows, MAS-MobileAgent achieves an MSRS of 0.613 on single-app tasks and 0.829 on cross-app tasks, substantially lower than the baseline MobileAgentV2's 1.058 and 1.247, respectively. This indicates that shortcut-augmented agents consistently execute paths closer to optimal solutions. Notably, MobileAgent-E, despite its low MS due to frequent premature termination from errors, maintains a competitive MSRS of 0.818 on single-app tasks, demonstrating that when it does succeed, its self-generated shortcuts enable efficient execution.

\paragraph{Agent-as-a-Model Performance Characteristics.}
The expanded Agent-as-a-Model results show a clear difference between general-purpose models and specialized GUI models. General-purpose models adapt more effectively to the hybrid action space: MAS-GLM-4.5V achieves the highest SR on both single-app (73.9\%) and cross-app tasks (57.4\%), while MAS-Qwen3-VL-32B and MAS-Qwen3-VL-235B obtain stable efficiency gains. In contrast, specialized GUI models do not automatically benefit from shortcuts. MAI-UI-8B improves with MAS, but ScaleCUA variants rarely invoke shortcuts and show marginal or negative changes. These results suggest that shortcut augmentation requires not only GUI perception but also reliable shortcut selection and tool-use reasoning.

\paragraph{Cost and Efficiency.}
Token cost analysis reveals that shortcut augmentation provides substantial efficiency gains while marginally increasing cost. On single-app tasks, MAS-MobileAgent reduces token cost by 17\% compared to MobileAgentV2, while on cross-app tasks, the reduction is 16\%. Mean execution time (MET) improvements are particularly pronounced on cross-app tasks: MAS-T3A reduces MET by 28\% compared to T3A (257s → 186s), demonstrating that shortcuts not only improve success rates but also significantly accelerate task completion through reduced step counts and more efficient execution paths.

\subsection{Details of MAS-MobileAgent and MAS-T3A}

Compared to MobileAgentV2, MAS-MobileAgent acquires the standard screenshot and UI elements and retrieves relevant and potentially applicable shortcuts from the shortcut knowledge base, based on the current task goal and interface context. This retrieved shortcut information is injected as an additional context into the prompt provided to the base model, along with screen information and the task goal. 

MAS-T3A operates on a similar principle, but its core distinction lies in its input modality: it relies on the structured UI tree, whereas MAS-MobileAgent processes visual screenshots. By evaluating both agents, we aim to demonstrate that the effectiveness of predefined shortcuts is framework-agnostic and can benefit agents with fundamentally different perceptual modalities.

\subsection{Shortcut Retrieval, Injection, and Action Space Mapping}
\label{app:shortcut_retrieval}

\paragraph{Retrieval and Injection Mechanism.} 
To enable efficient shortcut utilization, we implement a keyword-based retrieval mechanism. For each task instruction, we extract the mentioned application names and filter the shortcut knowledge base to retrieve only applicable shortcuts for those applications. The retrieved shortcuts are then injected into the agent's prompt as structured text descriptions with function signatures, parameters, and natural language descriptions (e.g., \texttt{Amazon.search\_product(query: str) - Searches for products matching the query in Amazon}). This application-scoped filtering reduces context length while ensuring all relevant shortcuts remain available. This controlled retrieval setting isolates shortcut selection and execution from open-domain retrieval failures. Future extensions can replace this module with unfiltered shortcut search or RAG-based shortcut retrieval, while our interference study (Appendix~\ref{app:shortcut_invocation_robustness}) provides an initial stress test under noisy shortcut pools.

\paragraph{Action Space Standardization.} 
To ensure fair evaluation across different shortcut types, we adopt the T3A action space as our standardization target. All shortcuts, whether predefined (APIs, deep-links, RPA scripts) or agent-generated, are mapped to T3A-compatible action sequences during execution. This unified framework enables direct comparison between different shortcut generation strategies.

\paragraph{Coordinate-Free Dynamic Shortcuts vs. Action Replay.} 
A critical design choice distinguishes agent-generated shortcuts. \textit{Dynamic shortcuts} specify UI elements through semantic descriptions (e.g., ``the search button'') rather than hardcoded coordinates or indices, requiring the agent to perform real-time grounding during execution. This design offloads the \textit{planning} burden to the shortcut itself while requiring only \textit{grounding} at runtime, reducing strategic errors and enhancing robustness to UI changes.

In contrast, \textit{action replay shortcuts} (task-level and subtask-level) record fixed sequences with specific coordinates and element indices. While simpler to generate, they are brittle to UI changes. As shown in Table~\ref{tab:generation_result}, this design difference significantly impacts Shortcut Success Rate (SSR): predefined shortcuts achieve 100\% SSR, dynamic shortcuts achieve 75\% SSR, while replay-task shortcuts achieve only 10\% SSR due to their sensitivity to UI variations. The subtask-level replay (SSR=73\%) performs better than task-level replay by abstracting shorter, more reusable patterns, but still suffers from coordinate brittleness. These results validate that coordinate-free dynamic shortcuts better balance generation complexity with execution robustness, making them more suitable for real-world deployment where application UIs evolve over time.

\subsection{Shortcut Invocation Robustness}
\label{app:shortcut_invocation_robustness}

To evaluate whether agents can use shortcuts reliably, we annotate the ground-truth shortcuts for each task and compare them with the shortcuts selected by each agent. Table~\ref{tab:shortcut_selection_quality} reports Mean Task Shortcut Recall (MTSR), Mean Shortcut Selection Precision (MSSP), MF1, and redundancy rate. The results show that shortcut selection quality is closely tied to efficiency gains. For example, GLM-4.5V maintains high precision with a low redundancy rate (4.8\%), leading to an 18.5\% reduction in MS. In contrast, Qwen3-VL-4B has lower precision (0.439) and higher redundancy (33.5\%), causing MS to increase by 16.3\% despite a small SR gain.

\begin{table*}[t]
\centering
\resizebox{0.92\textwidth}{!}{%
\begin{tabular}{lcccccc}
\toprule
Agent & MTSR$\uparrow$ & MSSP$\uparrow$ & MF1$\uparrow$ & Redun.(\%)$\downarrow$ & SR Impr.(\%)$\uparrow$ & MS Red.(\%)$\uparrow$ \\
\midrule
MAS-MobileAgent & \underline{0.709} & \textbf{0.721} & \textbf{0.715} & \underline{13.866} & \textbf{+79.452} & \textbf{+32.847} \\
MAS-Qwen3-VL-32B & \textbf{0.721} & 0.540 & \underline{0.617} & 44.784 & +31.803 & +10.771 \\
MAS-GLM-4.5V & 0.555 & 0.681 & 0.607 & \textbf{4.848} & +29.996 & +18.489 \\
MAS-Qwen3-VL-235B & 0.674 & 0.550 & 0.605 & 31.156 & +25.930 & +13.128 \\
MAS-Qwen3-VL-8B & 0.496 & 0.601 & 0.539 & 23.404 & \underline{+63.914} & +9.985 \\
MAS-T3A & 0.457 & 0.632 & 0.530 & 22.500 & +22.252 & \underline{+18.963} \\
MAS-Qwen3-VL-4B & 0.422 & 0.439 & 0.428 & 33.498 & +4.162 & -16.277 \\
MAS-MAI-UI-8B & 0.218 & \underline{0.719} & 0.335 & 22.807 & +19.177 & +6.096 \\
MAS-ScaleCUA-32B & 0.000 & 0.000 & 0.000 & 0.000 & -6.383 & +1.424 \\
MAS-ScaleCUA-7B & 0.000 & 0.000 & 0.000 & 0.000 & +6.748 & -6.882 \\
\bottomrule
\end{tabular}
}
\caption{\textbf{Shortcut selection quality and performance changes on MAS-Bench.} Bold and underlined values denote the best and second-best results, respectively. MTSR: Mean Task Shortcut Recall; MSSP: Mean Shortcut Selection Precision; MF1: mean F1 score; Redun.: redundancy rate of shortcut calls, ranked among agents with nonzero shortcut use. SR Impr. and MS Red. denote relative success-rate improvement and mean-step reduction over the corresponding GUI-only baseline; negative MS Red. indicates increased mean steps.}
\label{tab:shortcut_selection_quality}
\end{table*}

We further test robustness in a no-relevant-shortcut setting. Specifically, we select 23 cross-app tasks with reasonable baseline success rates and remove all ground-truth shortcuts for each task while retaining task-irrelevant shortcuts as distractors. As shown in Table~\ref{tab:shortcut_interference}, incorrect shortcut invocation consistently hurts performance. Qwen3-VL-4B makes the most irrelevant shortcut calls (MISC=2.478) and suffers the largest SR drop ($-66.67\%$), whereas GLM-4.5V makes fewer irrelevant calls (MISC=0.478) and remains close to its GUI-only baseline ($-7.69\%$ SR). These results show that MAS-Bench can expose and penalize shortcut misuse, making it suitable for evaluating agents under larger and noisier shortcut pools.

\begin{table*}[t]
\centering
\resizebox{0.82\textwidth}{!}{%
\begin{tabular}{lccccc}
\toprule
Agent & BL SR$\uparrow$ & $\Delta$SR$\uparrow$ & BL MS$\downarrow$ & $\Delta$MS$\downarrow$ & MISC$\downarrow$ \\
\midrule
GLM-4.5V & \textbf{0.565} & \textbf{-0.043 (-7.69\%)} & 21.739 & \textbf{+0.087} & \textbf{0.478} \\
Qwen3-VL-235B & \underline{0.522} & \underline{-0.043 (-8.33\%)} & \underline{20.870} & \underline{+2.391} & 1.913 \\
Qwen3-VL-32B & 0.435 & -0.043 (-10.00\%) & \textbf{20.391} & +4.783 & \underline{1.087} \\
Qwen3-VL-8B & 0.348 & -0.087 (-25.00\%) & 21.000 & +5.348 & 1.261 \\
Qwen3-VL-4B & 0.391 & -0.261 (-66.67\%) & 22.261 & +9.652 & 2.478 \\
\bottomrule
\end{tabular}
}
\caption{\textbf{Resistance to task-irrelevant shortcut interference on 23 cross-app tasks.} Bold and underlined values denote the best and second-best results, respectively. BL denotes the GUI-only baseline. In the interference setting, all ground-truth shortcuts are removed and only task-irrelevant shortcuts are exposed. MISC denotes the mean number of irrelevant shortcut calls per task. For $\Delta$SR, larger values indicate smaller degradation, with the relative drop in parentheses used to break ties.}
\label{tab:shortcut_interference}
\end{table*}

\subsection{Success Rate on Different-Level Tasks}
\label{app:different_level_tasks}

As shown in Table~\ref{table:mean_steps}, an analysis of metrics such as MSRS reveals that the GUI-shortcut hybrid agent, by utilizing the predefined shortcuts knowledge base, demonstrates improvements in both success rate and efficiency across tasks of varying difficulty levels when compared to the GUI-only agent.

\begin{table}[h!]
\centering
\resizebox{0.99\linewidth}{!}{%
\begin{tabular}{l|c|>{\centering\arraybackslash}p{0.65cm}>{\centering\arraybackslash}p{0.65cm}>{\centering\arraybackslash}p{0.8cm}}
\toprule % 顶线
\textbf{Agent} & \textbf{DL} & \textbf{SR} & \textbf{MS} & \textbf{MSRS} \\
\midrule % 中间线
\multirow{3}{*}{T3A} & Level 1 & 71.05 & 7.53 & 0.94 \\
 & Level 2 & 36.36 & 17.62 & 1.08 \\ % MSR 修正为两位小数以对齐
 & Level 3 & 45.71 & 21.17 & 1.08 \\
\midrule
\multirow{3}{*}{M3A} & Level 1 & 65.79 & 7.66 & 1.02 \\
 & Level 2 & 51.52 & 16.98 & 1.21 \\
 & Level 3 & 31.43 & 25.34 & 1.04 \\
\midrule
\multirow{3}{*}{MobileAgentV2} & Level 1 & 52.63 & 7.87 & 1.04 \\
 & Level 2 & 34.85 & 20.05 & 1.14 \\
 & Level 3 & 17.14 & 30.11 & 1.35 \\
\midrule
\multirow{3}{*}{MobileAgentE} & Level 1 & 44.74 & 3.66 & 0.83 \\
 & Level 2 & 20.00 & 5.15 & 0.97 \\
 & Level 3 & 17.14 & 5.43 & 0.60 \\
\midrule
\multirow{3}{*}{MAS-T3A} & Level 1 & 76.32 & 6.16 & 0.76 \\
 & Level 2 & 53.03 & 13.21 & 0.82 \\
 & Level 3 & 37.14 & 20.69 & 0.99 \\
\midrule
\multirow{3}{*}{MAS-MobileAgent} & Level 1 & 84.21 & 3.97 & 0.54 \\ % MS 修正为一位小数以对齐
 & Level 2 & 65.15 & 13.14 & 0.76 \\ % MSR 修正为两位小数以对齐
 & Level 3 & 37.14 & 22.26 & 0.77 \\
\bottomrule
\end{tabular}
}
\caption{\textbf{Performance comparison of Agentic Workflow agents (Gemini-2.5-Pro) on MAS-Bench across different difficulty levels.} Column definitions: \# DL (difficulty level), \# SR (success rate), \# MS (Mean step), \# MSRS (Mean Step Ratio on Successful tasks).}
\label{table:mean_steps}
\end{table}

Similarly, Table~\ref{table:mean_steps_model} presents the performance comparison of Agent-as-a-Model approaches across different difficulty levels. The results show that MAS-Qwen3-VL-8B consistently outperforms its GUI-only counterpart across all difficulty levels, with particularly notable improvements in MSRS, demonstrating the effectiveness of shortcut augmentation for this category of agents.

\begin{table}[h!]
\centering
\resizebox{0.99\linewidth}{!}{%
\begin{tabular}{l|c|>{\centering\arraybackslash}p{0.65cm}>{\centering\arraybackslash}p{0.65cm}>{\centering\arraybackslash}p{0.8cm}}
\toprule % 顶线
\textbf{Agent} & \textbf{DL} & \textbf{SR} & \textbf{MS} & \textbf{MSRS} \\
\midrule
\multirow{3}{*}{Qwen2.5-VL-3B} & Level 1 & 10.53 & 11.92 & 1.40 \\
 & Level 2 & 1.52 & 22.68 & 1.09 \\
 & Level 3 & 0.00 & 30.17 & - \\
\midrule
\multirow{3}{*}{Qwen2.5-VL-7B} & Level 1 & 7.89 & 11.95 & 1.50 \\
 & Level 2 & 0.00 & 23.08 & - \\
 & Level 3 & 0.00 & 30.23 & - \\
\midrule
\multirow{3}{*}{UI-TARS-1.5-7B} & Level 1 & 55.26 & 8.82 & 1.17 \\
 & Level 2 & 22.73 & 20.00 & 1.31 \\
 & Level 3 & 11.43 & 29.00 & 1.10 \\
\midrule
\multirow{3}{*}{GUI-Owl-7B} & Level 1 & 55.26 & 9.08 & 1.38 \\
 & Level 2 & 21.21 & 15.68 & 1.02 \\
 & Level 3 & 17.14 & 22.40 & 1.07 \\
\midrule
\multirow{3}{*}{Qwen3-VL-4B} & Level 1 & 44.74 & 9.61 & 1.26 \\
 & Level 2 & 13.64 & 19.33 & 1.39 \\
 & Level 3 & 17.14 & 22.80 & 0.88 \\
\midrule
\multirow{3}{*}{Qwen3-VL-8B} & Level 1 & 50.00 & 8.00 & 1.14 \\
 & Level 2 & 16.67 & 18.11 & 1.13 \\
 & Level 3 & 17.14 & 23.00 & 1.01 \\
\midrule
\multirow{3}{*}{MAS-Qwen3-VL-8B} & Level 1 & 63.16 & 7.16 & 0.85 \\
 & Level 2 & 37.88 & 15.98 & 0.89 \\
 & Level 3 & 28.57 & 21.34 & 0.71 \\
\bottomrule
\end{tabular}
}
\caption{\textbf{Performance comparison of Agent-as-a-Model approaches on MAS-Bench across different difficulty levels.} Column definitions: \# DL (difficulty level), \# SR (success rate), \# MS (Mean step), \# MSRS (Mean Step Ratio on Successful tasks).}
\label{table:mean_steps_model}
\end{table}

\subsection{Influence of Base Model Capability}
\label{app:diff_pro_flash}

To assess the impact of GUI-shortcut hybrid actions on base models with varying capabilities, we conducted experiments using Gemini-2.0-Flash and Gemini-2.5-Pro. As shown in Table~\ref{table:flash_diff_pro}, shortcut augmentation improves both models, but the gains appear in different forms. On single-app tasks, Gemini-2.0-Flash improves from 25.0\% to 40.2\% SR and reduces MS by 29\%, while Gemini-2.5-Pro improves from 44.6\% to 64.1\% SR and reduces MS by 27\%.

The role of shortcuts is clearer on cross-app tasks. With Gemini-2.0-Flash, the GUI-only agent fails all cross-app tasks (0\% SR), whereas MAS-MobileAgent reaches 23.4\% SR and reduces MET by 57.6\%. With Gemini-2.5-Pro, MAS-MobileAgent improves SR from 17.0\% to 61.7\% and reduces MS by 37\%. These results indicate that shortcuts can make weak base models solve tasks that are otherwise infeasible, while also improving the reliability and efficiency of stronger base models.

The Qwen3-VL series in Table~\ref{tab:results-masbench} provides a more fine-grained view of this trend. MAS-Qwen3-VL-8B achieves the largest relative SR gain (+64.1\%), suggesting that mid-scale models can benefit substantially once they have sufficient reasoning ability to use shortcuts. In contrast, MAS-Qwen3-VL-4B yields only a small SR gain (+3.9\%) and increases MS by 16.3\%, indicating that shortcut misuse can offset the efficiency benefit for under-capable models. Larger models, such as Qwen3-VL-32B and Qwen3-VL-235B, achieve more stable improvements (+32.0\% and +25.9\% SR), but their relative gains are smaller due to stronger GUI-only baselines. Overall, shortcut augmentation is most effective when the base model is capable enough to select shortcuts reliably but still benefits from bypassing long GUI interaction paths.

\section{Details of Shortcut Generation Evaluation}

\subsection{Shortcut Generation Method}
\label{app:shortcut_generation}

To validate the effectiveness of the MAS-Bench framework in evaluating an agent's shortcut generation capabilities, we generated a diverse set of agent-generated shortcuts using various methods. 

We employ M3A as our baseline agent to generate different types of shortcuts. Without altering its core operational logic, M3A autonomously explores the environment and generates task execution trajectories, utilizing Gemini-2.5-Pro as its base model. Upon completing each task, we construct a prompt containing the task instruction, the whole action history, and the reasoning for each step. This prompt is then provided to Gemini-2.5-Pro, generating the corresponding shortcuts.

\begin{figure}[t]
    \centering
    \includegraphics[width=\columnwidth]{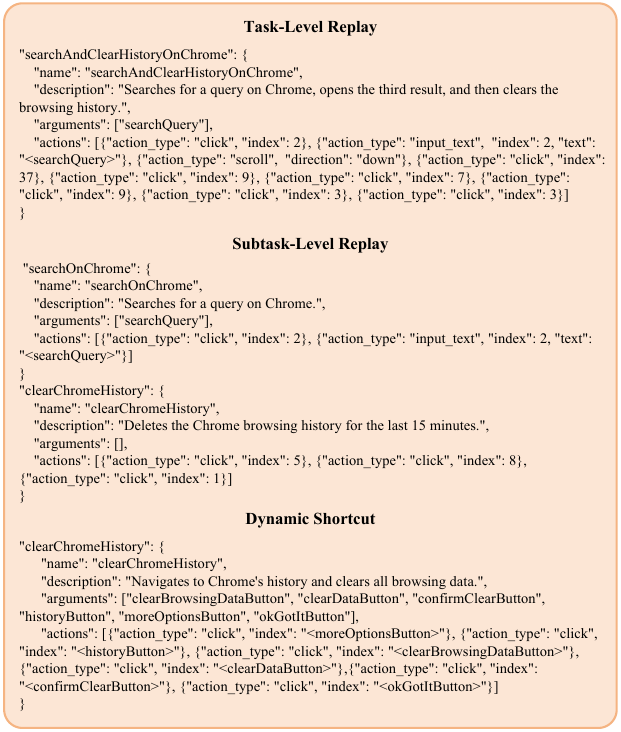}
    \caption{Examples of the resulting shortcut types. Action Replay shortcuts (Task-Level and Subtask-Level) use a sequence of actions with fixed indices, while Dynamic Shortcuts use variable arguments that correspond to UI elements.}
    \label{fig:shortcut_example}
\end{figure}

Examples of the resulting shortcut types are shown in Fig.~\ref{fig:shortcut_example}. As detailed in Table~\ref{table:shortcut_number}, this process resulted in 46 Subtask-level Macro-Replay, 45 Dynamic, 39 Task-level Macro-Replay, and 36 MobileAgent-E shortcuts.

\begin{table}[h]
\centering
\resizebox{0.99\linewidth}{!}{%
\begin{tabular}{lc}
\toprule
\textbf{Shortcut Type} & \textbf{Number of Shortcuts} \\
\midrule
$S_{\text{Replay-Task}}$   & 39 \\
$S_{\text{Replay-Subtask}}$ & 46 \\
$S_{\text{Dynamic}}$         & 45 \\
$S_{\text{MobileAgent-E}}$      & 36 \\
\bottomrule
\end{tabular}
}
\caption{Number of agent-generated shortcuts by different methods.}
\label{table:shortcut_number}
\end{table}

\subsection{Subset Task}
\label{app:subset_task_list}

To validate our framework for evaluating an agent's shortcut generation capability, we randomly selected a subset of 50\% from the 139 tasks in MAS-Bench. This subset comprises a total of 69 tasks, including 46 single-app and 23 cross-app tasks.

\section{Reliability Analysis of Automated Evaluation}
\label{app:eval_reliability}

\subsection{Evaluation Pipeline Architecture}

\begin{figure*}[t]
    \centering
    \includegraphics[width=\textwidth]{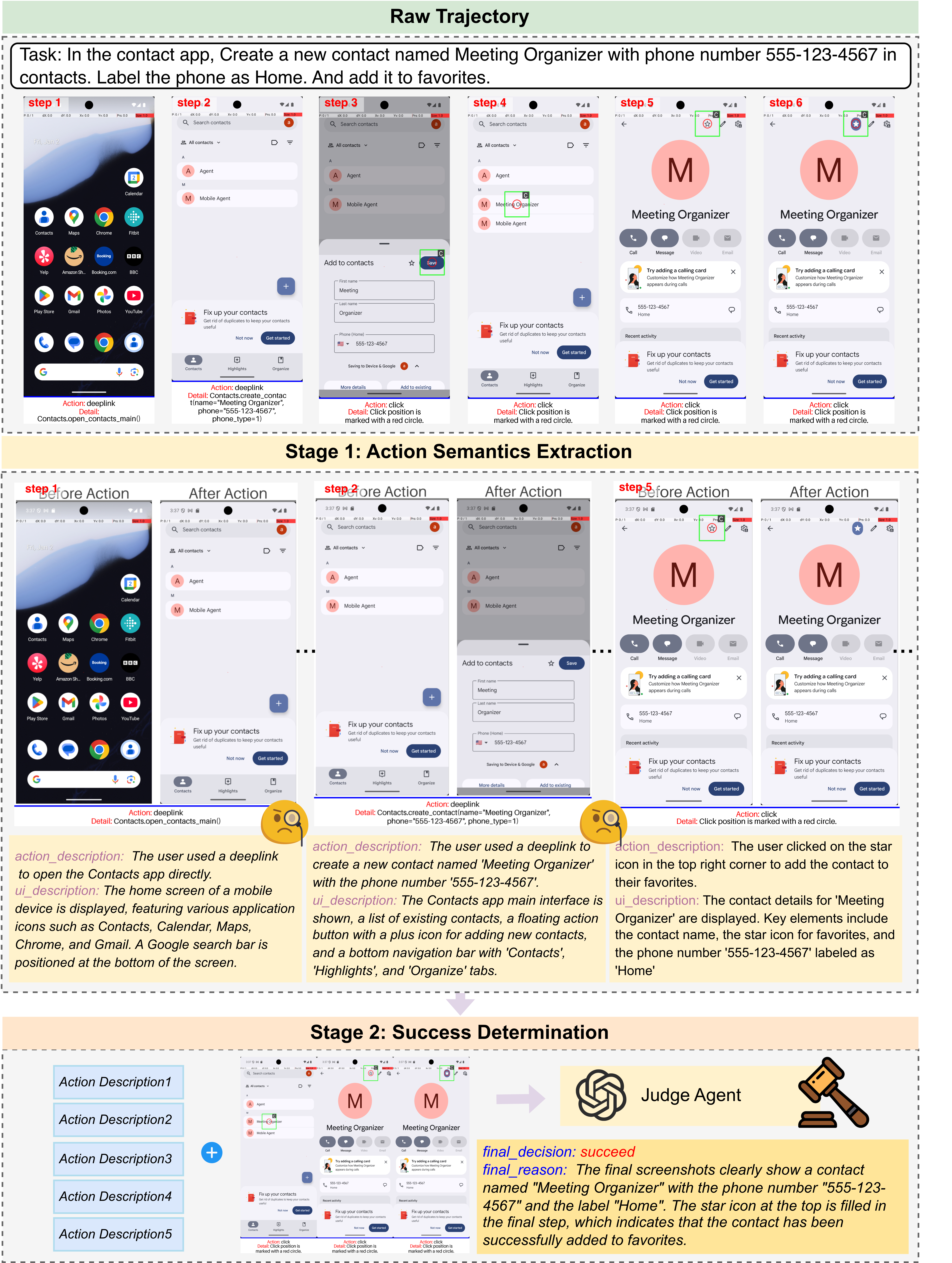}
    \caption{\textbf{Two-Stage Evaluation Pipeline of MAS-Bench-Eval.} Stage 1 (Action Semantics Extraction) employs an efficient MLLM to generate step-by-step textual descriptions of executed actions and UI state transitions. Stage 2 (Success Determination) uses a highly capable MLLM to judge task completion by reasoning over the task instruction, action descriptions from Stage 1, and the final three screenshots.}
    \label{fig:evaluation_pipeline}
\end{figure*}

Fig.~\ref{fig:evaluation_pipeline} illustrates the two-stage \textit{Describe-and-Judge} framework underlying MAS-Bench-Eval. This decoupled architecture optimizes the trade-off between computational efficiency and evaluation accuracy.

\paragraph{Stage 1: Action Semantics Extraction.} 
An efficient MLLM processes the agent's execution trajectory step-by-step, analyzing each screenshot-action pair to generate concise textual captions. Each caption describes both the executed action (e.g., ``invoked search\_product shortcut'') and the resulting UI state transition (e.g., ``navigated to search results page''). This stage transforms potentially lengthy visual trajectories into structured semantic histories, enabling efficient downstream processing while preserving critical behavioral information.

\paragraph{Stage 2: Success Determination.} 
A highly capable MLLM serves as the judge, receiving three inputs: the task instruction, the step-by-step action descriptions from Stage 1, and the final three screenshots. The judge performs compositional reasoning to verify whether all task requirements are satisfied, including sub-goal completion, final state correctness, and workflow integrity. It outputs a binary verdict with justification.

This decoupled architecture provides three key advantages: (1) \textit{Efficiency}: lightweight models handle repetitive trajectory processing, while capable models focus on critical judgment; (2) \textit{Interpretability}: semantic histories enable transparent evaluation and failure analysis; (3) \textit{Modularity}: individual components can be upgraded independently as stronger models emerge.

\subsection{Validation Against Human Judgment}

To validate the reliability of \textbf{MAS-Bench-Eval}, we conducted a comprehensive study comparing its automated verdicts against manual verification. We sampled 278 execution trajectories generated by MAS-MobileAgent on MAS-Bench, covering a diverse set of 184 single-app and 94 cross-app tasks. All trajectories were manually annotated to establish ground truth labels for task success. In our implementation of MAS-Bench-Eval, we instantiated the efficient VLM in \textbf{Stage 1 (Action Semantics Extraction)} with \texttt{Gemini-2.5-Flash} to generate step-by-step trajectory captions. For \textbf{Stage 2 (Success Determination)}, we employed the highly capable \texttt{Gemini-2.5-Pro} as the judge. This model reasons over the task instruction, the extracted semantic history, and the final three screenshots to issue a final success verdict.

Table~\ref{table:eval_reliability} reports the quantitative alignment between MAS-Bench-Eval and human evaluation. The results demonstrate remarkable consistency, with F1 scores exceeding 95\% across both scenarios (96.3\% for single-app and 95.1\% for cross-app). Specifically, the high precision (up to 98.1\%) indicates a minimal false-positive rate, ensuring that the benchmark does not overestimate agent performance. Meanwhile, the strong recall (up to 98.0\%) confirms the system's sensitivity in identifying true successes. These metrics verify that MAS-Bench-Eval serves as a reliable and scalable proxy for human assessment, enabling extensive experiments without the need for manual intervention.

% Evaluation reliability metrics for MAS-Bench-Eval
\begin{table}[t]
\centering
\resizebox{0.99\linewidth}{!}{%
\begin{tabular}{lccc}
\toprule
\textbf{Task Type} & \textbf{F1} & \textbf{Precision} & \textbf{Recall} \\
\midrule
Single App (184 tasks) & 96.3 & 98.1 & 94.5 \\
Cross App (94 tasks) & 95.1 & 92.5 & 98.0 \\
\bottomrule
\end{tabular}
}
\caption{\textbf{Reliability of MAS-Bench-Eval.} We evaluate the automated evaluation system against manual verification on 278 tasks.}
\label{table:eval_reliability}
\end{table}

\section{Bad Case Analysis}
We further analyze the performance of the GUI-shortcut hybrid agent on failed tasks, with a particular focus on cases where the GUI-only agent succeeds, but the GUI-shortcut hybrid agent fails. Our analysis of these failure cases reveals three primary.

\begin{figure*}[h]
    \centering
    \includegraphics[width=\textwidth]{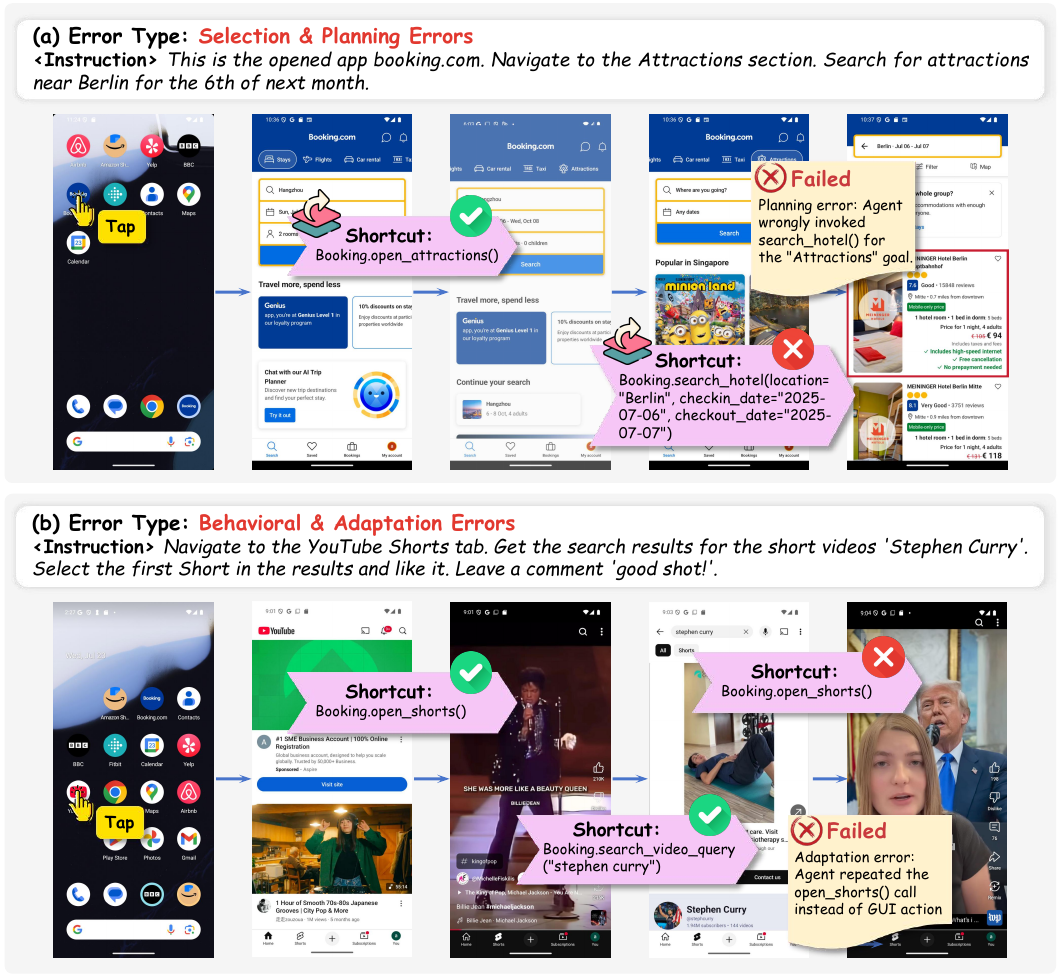}
    \caption{\textbf{Examples of GUI-shortcut hybrid agent failure cases.} (a) Selection and Planning Error: The agent incorrectly invokes search\_hotel() shortcut instead of searching for attractions, demonstrating failure in mapping natural language instructions to appropriate shortcuts. (b) Behavioral and Adaptation Error: The agent repeatedly calls open\_shorts() shortcut instead of switching to a GUI-only action to select the video, showing an inability to adapt operational mode based on task state.}
    \label{fig:badcase}
\end{figure*}

\paragraph{Selection and Planning Errors.} This category of error pertains to flaws in the agent's task comprehension and strategic planning. These errors typically occur when the agent fails to accurately map the natural language instruction to the correct shortcut, resulting in the invocation of an irrelevant shortcut that leads to task failure.

For instance, in an experiment with MAS-MobileAgent using Gemini-2.0-Flash (Fig.~\ref{fig:badcase}(a)), the agent was given the instruction: ``Navigate to the Attractions section. Search for attractions near Berlin for the 6th of next month.'' After launching the application, the agent made a planning error by incorrectly invoking the \texttt{search\_hotel()} shortcut, which was irrelevant to the ``Attractions'' goal, thus causing the task to fail.

\paragraph{Behavioral and Adaptation Errors.} This error class occurs when the agent fails to evaluate the outcome of an action and adjust its subsequent action. The deficiency typically stems from an inability to learn from the action histories and modify future behavior accordingly.

As shown in Fig.~\ref{fig:badcase}(b), when tasked to ``Get the search results for the short videos Stephen Curry. Select the first Short in the results and like it. Leave a comment good shot!'', The agent successfully uses shortcuts to search for the video. However, at the step requiring it to select the first video, it fails to switch to a GUI-based action and instead erroneously calls the \texttt{open\_shorts()} shortcut again. This results in an unproductive loop, demonstrating a failure to adapt its operational mode based on the task's state.

\paragraph{Execution and Formulation Errors.} This type of error is specific to the invocation of agent-generated shortcuts. As noted in the main text, the execution success rate for these shortcuts is considerably low. Failures typically result from the shortcut being poorly formulated during the generation phase, rendering it inherently flawed or not robust enough to handle slight variations in the UI during execution.

\begin{table*}[!t]
    \centering
    \resizebox{0.99\textwidth}{!}{%
    \begin{tabular}{l>{\centering\arraybackslash}p{1.2cm}>{\centering\arraybackslash}p{1.2cm}p{10cm}}
    \toprule
    \textbf{App name} & \textbf{Tasks} & \textbf{Shortcuts} & \textbf{Shortcut Example} \\
    \midrule
    YouTube & 11 & 5 & \texttt{youtube.search\_video\_query("cat")}\newline\texttt{youtube.open\_shorts()} \\

    Amazon & 10 & 10 & \texttt{amazon.open\_amazon()}\newline\texttt{amazon.open\_cart()} \\
    
    Booking.com & 10 & 10 & \texttt{booking.open\_hotelid()}\newline\texttt{booking.open\_my\_trips()} \\

    Google Calendar & 9 & 2 & \texttt{calendar.open\_calendar\_main()}\newline\texttt{calendar.create\_event(title="Meeting")} \\
    
    Google Maps & 9 & 8 & \texttt{google\_map.navigate\_to\_address("Beijing")}\newline\texttt{google\_map.search\_address("Square")} \\
    
    BBC & 8 & 16 & \texttt{bbc.open\_news()}\newline\texttt{bbc.open\_live()} \\

    Fitbit & 8 & 21 & \texttt{fitbit.open\_fitbit\_main()}\newline\texttt{fitbit.open\_sleep\_log()} \\

    Gmail & 8 & 3 & \texttt{gmail.open\_gmail()}\newline\texttt{gmail.open\_setting()} \\

    Chrome & 7 & 4 & \texttt{chrome.search\_query("Python")}\newline\texttt{chrome.open\_incognito()} \\

    Yelp & 8 & 6 & \texttt{yelp.open\_yelp\_main()}\newline\texttt{yelp.open\_events()} \\

    Contacts & 4 & 3 & \texttt{contacts.open\_contacts\_main()}\newline\texttt{contacts.open\_contact\_starred()} \\
    \bottomrule
    \end{tabular}
    }
    \caption{List of MAS-Bench apps and number of single-app tasks and shortcuts for each one.}
    \label{table:app-shortcut}
\end{table*}

\end{document}